%% file: main.tex
\documentclass{article}

\usepackage{iclr2027_conference, times}
\iclrfinalcopy
\input{math_commands}

\usepackage[utf8]{inputenc}
\usepackage[T1]{fontenc}
\usepackage{hyperref}
\usepackage{url}
\usepackage{booktabs}
\usepackage{amsfonts}
\usepackage{nicefrac}
\usepackage{microtype}
\usepackage{xcolor}
\usepackage{mathtools}
\usepackage{amsmath,amssymb}
\usepackage{enumitem}
\usepackage{algorithm}
\usepackage{algpseudocode}
\usepackage{graphicx}
\usepackage{comment} 
\usepackage{wrapfig} 
\usepackage{multirow} 
\usepackage{caption} 

\newcommand{\hmdlm}{\textsc{HMDLM}}
\newcommand{\mdlm}{\textsc{MDLM}}
\newcommand{\llada}{LLaDA}  
\newcommand{\mask}{\texttt{[MASK]}}

\title{Hierarchical Masked Diffusion Language \\ Models}

\author{
Parul Gupta$^{1}$ \quad
Satwik$^{1}$ \quad
Avni Aggarwal$^{1}$\thanks{Equal contribution.} \quad
Bhavesh Gurnani$^{1}$\footnotemark[1] \quad \\
\textbf{Arnab Kumar Mondal$^{2}$ \quad 
Dinesh Khandelwal$^{2}$ \quad
Dinesh Garg$^{2}$ \quad
Parag Singla$^{2}$}\quad \\
$^{1}$Indian Institute of Technology Delhi \quad $^{2}$IBM Research \\
}
\begin{document}
\maketitle
\lhead{Preprint} 

\input{abstract}
\input{introduction}
\input{related_work}
\input{background}

\input{hmdlm}

\input{experiments_orig}

\input{conclusion}

\clearpage
\bibliographystyle{iclr2027_conference}
\bibliography{references}
\newpage
\appendix
\input{appendix}

\end{document}

%% file: math_commands.tex
\usepackage{amsmath,amsfonts,bm}

\def\eqref#1{equation~\ref{#1}}

\def\1{\bm{1}}

\DeclareMathAlphabet{\mathsfit}{\encodingdefault}{\sfdefault}{m}{sl}
\SetMathAlphabet{\mathsfit}{bold}{\encodingdefault}{\sfdefault}{bx}{n}



%% file: abstract.tex
\begin{abstract}
Masked diffusion language models have emerged as an alternative to autoregressive models, enabling parallel token generation. However, standard masked diffusion language models use a shared noise schedule across positions without explicitly encoding which positions should be resolved first. Many tasks exhibit a natural priority among positions: document titles and section headings can guide the generation of body text, while Sudoku cells determined directly from the initial clues can help resolve more difficult cells. To incorporate this notion of priority (or hierarchy) among token positions, we introduce the Hierarchical Masked Diffusion Language Model (HMDLM), which augments each token position with a hierarchy level available as part of the training data, and requires zero or minimal additional annotations. It factors generation into two passes: the first predicts a hierarchy distribution over masked positions, the second predicts token distributions conditioned on it. Giving each hierarchy level its own noise schedule induces a soft, controllable preference for revealing higher-priority positions earlier while keeping decoding fully parallel. Our formulation is mathematically grounded, and we re-derive the forward and reverse processes and the NELBO for this factorization. Our evaluation shows that for the task of language generation on LM1B, HMDLM reduces perplexity by more than 1.5 points over Anchored Diffusion Language Model (ADLM), a recently proposed strong diffusion baseline; on WikiHow, we see a drop in perplexity 
of upto 4.4 pts on ADLM and upto 7 pts against MDLM (Masked Diffusion Language Model). A LLM judge on WikiHow, prefers HMDLM generations (the two variants combined) more than $2\times$ as often as the baseline. 
On the task of solving Sudoku puzzles requiring multi-step reasoning, it improves solution accuracy by up to $8.6\%$ absolute points over existing diffusion baselines.  
\end{abstract}

%% file: introduction.tex
    \section{Introduction}
\label{sec:intro}

Large language models (LLMs) have revolutionized the AI space, from general-purpose text generation to solving complex problems such as mathematical reasoning and theorem proving. Despite their success, limitations remain. One of the criticisms of modern-day LLMs, which are largely auto-regressive in nature, is that they generate text only in a strictly left-to-right manner. It has been argued that this style of decoding has specific disadvantages, such as not being able to correct past mistakes, and involving large computational costs. As a remedy, recent literature has proposed another class of models, referred to as {\em Diffusion Models}~\citep{austin2021structured,lou2024discrete,sahoo2024simple,nie2025largelanguagediffusionmodels}, which do not subscribe to the left-to-right generation style, but rather generate multiple tokens in parallel. The formulation is based on a forward diffusion process, which translates the text into noise, and on learning the reverse process, can generate the desired text by iteratively denoising from complete noise. This line of work has shown significant promise for the future of language generation, both for increasing computational efficiency and solving complex tasks (e.g., solving Sudoku or generating structured text).

A key open question in masked diffusion language models is how to choose the order in which positions should be denoised. Standard masked diffusion treats token positions symmetrically: with a shared masking schedule and no task-dependent notion of priority. Recent work shows that token ordering and decoding strategies can substantially affect generation quality \citep{kim2025train,garg2025masked, hong2026unifying,fox2026learning}. Recent approaches address this by identifying important anchor tokens \citep{rout2025anchoreddiffusionlanguagemodel, xue2026ancoder}, or adding planning-aware training and decoding mechanisms \citep{peng2025planner,israel2025planned}. These methods suggest that not all positions should be treated equally, but none provides a general hierarchy-aware forward and backward diffusion process. Such priority is often available: many problems carry a natural hierarchy over positions. In document generation, titles and section headings or summary often provide high-level semantic structure before the surrounding body text. In Sudoku, more constrained cells are easier to fill first, and their priority belongs to the position, not the digit it will contain.

We introduce \hmdlm{}, a hierarchical extension of masked diffusion language models  that associates each token position with a hierarchy variable. 
During the forward process, both the token and its hierarchy state are diffused toward a masked state. During the reverse process, the model first predicts the hierarchy for masked positions and then predicts token identities conditioned on the predicted hierarchy. This two-stage factorization allows the model to learn which hierarchy level each position belongs to before decoding its token value.  Unlike anchor-based methods~\citep{rout2025anchoreddiffusionlanguagemodel}, which tie importance to token identities, \hmdlm{} assigns hierarchy to positions, with a separate noise schedule per level to control how fast each level is unmasked (see Section~\ref{sec:noise_sch}). We also derive the negative evidence lower bound (NELBO) for this process which serves as our training objective \ref{sec:hmdlm}.

We experimentally validate HMDLM across several benchmark domains. On language modeling benchmarks like LM1B, we train the model from scratch and compare with diffusion baselines (like  MDLM~\citep{sahoo2024simple}, SEDD~\citep{lou2024discrete}, the Anchored Diffusion Language Model (ADLM)~\citep{rout2025anchoreddiffusionlanguagemodel}, and DiffusionBERT~\citep{he2023diffusionbert}) on both perplexity and generation quality. On Wikihow, we fine-tune on pretrained checkpoints and compare with ADLM and MDLM on perplexity and qualitative generation. On Sudoku, we train from scratch, and more reliably produce accurate solutions than the diffusion baselines (like MGDM~\citep{ye2025beyond} and MDLM). These results demonstrate that explicitly modeling hierarchy over token positions can improve masked diffusion generation across both natural language and structured domains.

Our work makes the following contributions: (a) We present a novel formulation for incorporating the notion of hierarchy in the discrete diffusion models, referred to as HMDLM (b) We re-formulate the standard diffusion formulation to have a two stage training and inference process, where the first stage predicts distribution over hierarchy levels, and second one predicts token distribution given the hierarchy prediction.  (c) We re-derive the ELBO to incorporate the notion of hierarchy in the forward and backward diffusion process. (d) We present experiments showing gains over masked diffusion baselines in perplexity and generation quality on LM1B and WikiHow, and report accuracy on Sudoku. {We also plan to release the code for HMDLM on paper acceptance.}

%% file: related_work.tex
\section{Related Work}
\label{sec:related_work}

\paragraph{Masked diffusion language models:}
Diffusion models have recently been adapted to language generation as an alternative to left-to-right autoregressive modeling. Early approaches explored diffusion in continuous word-vector spaces, as in Diffusion-LM~\citep{li2022diffusion}, or directly over discrete tokens, as in D3PMs~\citep{austin2021structured} with structured transition matrices and absorbing \mask{} states. More recent masked diffusion language models formulate generation as iterative denoising from partially masked sequences: the forward process gradually replaces tokens with \mask{}, and the reverse model reconstructs the sequence from bidirectional context. SEDD~\citep{lou2024discrete}, reparameterized discrete diffusion~\citep{zheng2023reparameterized}, DiffusionBERT~\citep{he2023diffusionbert}, MD4~\citep{shi2024simplified} and \mdlm{}~\citep{sahoo2024simple} improve the objectives and sampling procedures for this setting, while large-scale systems such as MDM ~\citep{ICLR2025_ce1c1ff5}, \llada{}~\citep{nie2025large} and LLaDA2.0~\citep{bie2025llada2} show that masked diffusion can support instruction-following LLMs. However, these models typically employ a uniform masking schedule, ignoring the inherent semantic or structural hierarchy of the task. 

\paragraph{Generation order and decoding strategies:}
Masked diffusion is attractive because generation need not proceed left-to-right; tokens can be revealed in arbitrary or partially parallel orders. This flexibility also makes the choice of denoising order important. Prior work shows that adaptive or learned ordering can significantly improve masked diffusion, through inference-time token ordering \citep{kim2025train}, planner-aware training \citep{peng2025planner}, planned generation where the model first constructs a high-level plan and then denoises the planned spans in parallel \citep{israel2025planned}, variational order learning \citep{fox2026learning}, or unified learnable-order formulations \citep{hong2026unifying}. Other methods~\citep{wang2025remasking,huang2025pc} modify sampling trajectories to address irreversibility or decoding bias. Our work is complementary: rather than learning a free-form position policy or adding an external planner, \hmdlm{} encodes a task-dependent hierarchy in the diffusion state itself and uses per-level schedules to induce a controllable soft generation order.

\paragraph{Anchoring, planning, and structured priors:}
Several recent works also argue that diffusion language models benefit from identifying important tokens or high-level structure before completing the full sequence. Anchored Diffusion Language Models~\citep{rout2025anchoreddiffusionlanguagemodel} observe that masking important words early can remove useful context, and use a two-stage model that predicts anchor-token distributions before predicting missing tokens. Planned Diffusion~\citep{israel2025planned} first constructs an explicit plan and then generates spans in parallel. In code generation, AnCoder~\citep{xue2026ancoder} uses syntax-tree priors to prioritize keywords and identifiers, and code-focused diffusion models such as DiffuCoder and Dream-Coder further show the value of domain-specific structure~\citep{gong2025diffucoder,xie2025dream}. These works share our motivation that not all sequence positions should be treated equally. However, \hmdlm{} differs by attaching hierarchy to positions rather than only token identities, supporting multiple task-defined hierarchy levels, and using these levels to control both the forward corruption process and the reverse denoising dynamics through level-specific schedules.

%% file: background.tex
\section{Background}
\label{sec:background}

\subsection{Notations}
Let ${x}$ denote a sequence of discrete tokens each token $x^i$ is represented as one-hot column vector, and $\mathcal{V} = \{x^i \in \{0,1\}^D : \sum_{j=1}^D x^i[j] = 1\}$ at position $i$ as the set of all such vectors. Define $Cat(.;\pi)$ as the categorical distribution over $\mathcal{V}$ classes, where $\pi \in \mathbb{R}^D$ and $\sum_d \pi[d] = 1$. Also we assume a special $[MASK]$ token, $m \in  \mathcal{V}$ and treat it as the $D^{th}$ category, i.e., $m[D] = 1$. Let $x_t$ denote the state of the diffusion process at time step $t$. The following description follows the treatment provided in~\citet{sahoo2024simple}.

\subsection{Discrete Diffusion Models}

\textbf{Forward process.}
Discrete diffusion process is defined using a forward process that gradually corrupts data over $T$ time steps, and replaces each token with a special MASK token. Let  $t \in \{0, 1, \ldots, T\}$, with $t = 0$ corresponding to clean data and $t = T$ to fully masked data, and $x_t^i \in \mathcal{V}$ denote progressively noisier latent variables where the timestep $t$ ranges from $0$ (zero noise) to $T$ (maximum noise). The forward noising process $q$ at time $t$ and position $i$ is defined as:
\begin{equation}
q(x_{t}^i \mid x_0^i) = \mathrm{Cat}\!\left(x_t^i; \alpha_t x_0^i + (1 - \alpha_t)m\right),
\label{eq:mdlm_forward}
\end{equation}
where $x_0^i$ is the clean state, $\alpha_t$ is the masking schedule where $\alpha_t \in [0, 1]$ is a strictly decreasing function in $t$, with $\alpha_0 = 1$ and $\alpha_T = 0$ controlling the proportion of unmasked tokens at time $t$. The corresponding posterior transition with $\lambda_t = \frac{\alpha_{t-1}-\alpha_t}{1-\alpha_t}$ is:
\begin{equation}
q(x_{t-1}^i \mid x_t^i, x_0^i) =
\begin{cases}
\mathrm{Cat}(x_{t-1}^i; x_t^i), & x_t^i \neq m, \\
\mathrm{Cat}\!\left(x_{t-1}^i; (1-\lambda_t)m + \lambda_t x_0^i\right), & x_t^i = m.
\end{cases}
\label{eq:mdlm_posterior}
\end{equation}

\textbf{Reverse process.}
The reverse process iteratively denoises the data from timestep $t$ to $t-1$ until the sequence becomes fully unmasked. Diffusion model aims to approximate the reverse process by learning a backward model $p_\theta(x_{t-1}^i | x_t^i)$. Since the true posterior depends on the unknown clean token $x_0^i$, it is approximated with a neural network $x^i_\theta(x_t, t) \in \Delta^\mathcal{V}$, which predicts the clean token distribution from the noisy input $x_t^i$ at time $t$. During the reverse diffusion process, unmasked tokens stay fixed, and clean inputs are never masked:
\begin{equation}
p_{\theta}(x_{t-1}^i \mid x_t) =
\begin{cases}
\mathrm{Cat}(x_{t-1}^i; x_t^i), & x_t^i \neq m, \\
\mathrm{Cat}\!\left(x_{t-1}^i; (1-\lambda_t)m + \lambda_t\, x^i_{\theta}(x_t,t)\right), & x_t^i = m.
\end{cases}
\label{eq:mdlm_reverse}
\end{equation}

\textbf{NELBO.}
The denoising model $x_{\theta}({x_t}, t)$ is trained using the following Negative ELBO:
\begin{equation}
\begin{aligned} 
\mathcal{L}_{\mathrm{NELBO}}({x}; \theta) =
\mathbb{E}_{q,t}\!\left[
-\log p_{\theta}({x_0} \mid {x_1})
+ T \left(
\frac{\alpha_t - \alpha_{t-1}}{1 - \alpha_t}
\log \langle x_{\theta}(x_t, t), x_0 \rangle
\right)
\right]
\end{aligned}
\end{equation}

%% file: hmdlm.tex
\section{Hierarchical Masked Diffusion Language Model}
\label{sec:hmdlm}

In this section we introduce the Hierarchical Masked Diffusion Language Model
(HMDLM), which extends discrete diffusion models by incorporating a notion of
hierarchy into both the forward and backward processes. Our framework augments
each token position with a hierarchy level. Let $h_0^i$ denote the gold hierarchy at position $i$, represented as a one-hot vector over $K$ levels, i.e., $h_0^i \in \{0,1\}^K$. We introduce a distinct hierarchy mask state $h_{\mathrm{m}}$, giving $K{+}1$ possible hierarchy states. 
Positions to be unmasked at time step $t$ are governed by the level dependent masking schedule $\alpha_k(t)$ defined for time step $t \in \{0, \ldots, T\}$ and level $k \in \{0, \ldots, K-1\}$ which is a decreasing function. If a position $i$ at time step $t$ is sampled or is already unmasked, we then set $b_t^i=0$ and $b_t^i=1$ otherwise. \begin{equation}
s_t^i=(x_t^i,h_t^i)=(x_0^i,h_0^i)\ \text{if }b_t^i=0\ (\text{revealed}),\qquad (m,h_{\mathrm{m}})\ \text{if }b_t^i=1\ (\text{masked}).
\label{eq:forward_bernoulli}
\end{equation}
Since $b_t^i$ is shared, a token and its hierarchy level are always masked or
revealed together, ruling out decoupled states such as a masked token with a
revealed level.

\subsection{Forward and Reverse Process}
\textbf{Forward Process}
The forward process is a fixed Markov chain, that progressively masks tokens from the initial state $s_0$ to a noisy state $s_t$ at each time step $t$ over $T$ time steps, according to the hierarchy conditioned noise schedule $\alpha_k(t)$. Tokens at lower hierarchy levels are masked more aggressively at early time-steps, than tokens at higher hierarchy levels. The forward marginal at a single token position $i$ with hierarchy $k$ and joint state $s^i_{\mathrm{t}} = (x^i_{\mathrm{t}}, h^i_{\mathrm{t}})$ is given by:
\begin{equation}
q(s_t^i \mid s_{0}^i) = \text{Cat}\left(s_t^i;\, \alpha_{h_0^i}(t)\, s_{0}^i + (1-\alpha_{h_0^i}(t))\, {s_m}\right)
\label{eq:fwd_diff}
\end{equation}
At $t=T$, all tokens are fully masked regardless of the hierarchy level, i.e. $\alpha_k(T) = 0$. 
Let $\lambda_k(t)=\frac{\alpha_k(t-1)-\alpha_k(t)}{1-\alpha_k(t)}$ denote the probability that a masked level-$k$ position is revealed at step $t$, with $\lambda_k(t) = 0$ when $\alpha_k(t) = 1$. The true joint posterior is: 
\begin{equation}
q(s_{t-1}^i \mid s_t^i,s_0^i)=
\begin{cases}
\mathrm{Cat}(s_{t-1}^i;s_t^i), & s_t^i\neq\mathbf{s}_m,\\

\mathrm{Cat}\!\left(s_{t-1}^i;(1-\lambda_{h_0^i}(t))\mathbf{s}_m+\lambda_{h_0^i}(t)s_0^i\right), & s_t^i=\mathbf{s}_m.
\end{cases}
\label{eq:posterior_joint}
\end{equation}

\textbf{Reverse Process}

The reverse process aims to recover the clean sequence \(s_0\) from a fully masked sequence \(s_T\) over \(T\) denoising steps. We approximate the reverse process \(p_\theta(s_{t-1} \mid s_t, t)\) using the neural network \({s}_{\theta}(s_t,t)\), which is composed of two joined networks - the hierarchy head, $h_\theta(x_t,h_t,t)$ and the token head, $x_\theta(x_t,h_{t-1},t)$. It takes as input the noisy state \(s_t\)=($x_t$,$h_t$) at timestep \(t\) and predicts the clean joint state \(s_0\).

Revealed positions are carried over unchanged, $p_\theta(s_{t-1}^i \mid s_t^i)=\mathrm{Cat}(s_{t-1}^i;s_t^i)$ if $s_t^i\neq\mathbf{s}_m$. For a masked position ($s_t^i=\mathbf{s}_m$), the model first draws the hierarchy state and then the token conditioned on it:
\begin{equation}
\begin{aligned}
p_\theta\bigl(h_{t-1}^i = k \mid x_t,h_t\bigr) &= \lambda_k(t)\,g_k^i(t), && k\in\{0,\dots,K-1\},\\
p_\theta\bigl(h_{t-1}^i = h_{\mathrm{m}} \mid x_t,h_t\bigr) &= 1-\pi_i(t), && \\
p_\theta\bigl(x_{t-1}^i \mid x_t,h_{t-1}\bigr) &= \mathrm{Cat}\bigl(x_{t-1}^i;\,f^i(t)\bigr), && h_{t-1}^i\neq h_{\mathrm{m}},\\
p_\theta\bigl(x_{t-1}^i = m \mid x_t,h_{t-1}\bigr) &= 1, && h_{t-1}^i = h_{\mathrm{m}},
\end{aligned}
\label{eq:reverse_model}
\end{equation}

where $g^i(t)=\mathrm{softmax}\bigl(h_\theta^i(x_t,h_t,t)\bigr)$ is the hierarchy head's distribution over the $K$ levels, $f^i(t)=\mathrm{softmax}\bigl(x_\theta^i(x_t,h_{t-1},t)\bigr)$ is the token head's distribution over the vocabulary, and
$\pi_i(t)=\sum_{k=0}^{K-1}\lambda_k(t)\,g_k^i(t)=\mathbb{E}_{k\sim g^i(t)}\bigl[\lambda_k(t)\bigr]$, is the total probability that position $i$ is revealed at step $t$. Since the gold hierarchy $h_0^i$ is unknown in the reverse process, $\pi_i(t)$ replaces the true reveal probability $\lambda_{h_0^i}(t)$ of \eqref{eq:posterior_joint} by its expectation under the predicted hierarchy distribution. Because $\lambda_{k'}(t) \geq \lambda_{k}(t)$ for $k' > k$, a masked position is more likely to be revealed at a higher level than at a lower one even when $g^i(t)$ is uniform, which is what induces the soft generation order (Appendix~\ref{app:ordering}).

\subsection{ELBO} We derive the discrete time ELBO \ref{eq:joint_discrete_elbo1} and its continuous-time limit \ref{eq:total_loss} for the generative process $p_\theta(s_{0})$. The detailed derivation is given in the Appendix \ref{app:discrete_elbo}.
The discrete ELBO is given as:
\begin{equation}
\log p_\theta(\mathbf{s}_{0}) \;\geq\; \underbrace{\mathbb{E}_q\!\bigl[\log p_\theta(\mathbf{s}_0 \mid \mathbf{s}_1)\bigr]}_{L_0} - \sum_{t=2}^{T} \underbrace{\mathbb{E}_q\!\Bigl[D_{\mathrm{KL}}\bigl(q(\mathbf{s}_{t-1} \mid \mathbf{s}_t, \mathbf{s}_0) \;\|\; p_\theta(\mathbf{s}_{t-1} \mid \mathbf{s}_t)\bigr)\Bigr]}_{L_t}.
\label{eq:joint_discrete_elbo1}
\end{equation}
Each $L_t$ is a KL over the state $(\mathbf{s}_{t}) = (x_t, h_t)$.
By the chain rule of KL divergence, $L_t$ can be expanded as:
\begin{align}
  L_t&=
  \underbrace{D_{\mathrm{KL}}\bigl(
    q(\mathbf{h}_{t-1} \mid \mathbf{x}_t, \mathbf{h}_t, \mathbf{x}_0, \mathbf{h}_0)
    \;\|\;
    p_\theta(\mathbf{h}_{t-1} \mid \mathbf{x}_t, \mathbf{h}_t)
  \bigr)}_{\text{hierarchy KL (Pass 1)}}
  \nonumber\\[4pt]
  &\quad+
  \underbrace{\mathbb{E}_{q(\mathbf{h}_{t-1} \mid \mathbf{x}_t, \mathbf{h}_t, \mathbf{x}_0, \mathbf{h}_0)}\!\Bigl[
    D_{\mathrm{KL}}\bigl(
      q(\mathbf{x}_{t-1} \mid \mathbf{x}_t, \mathbf{h}_{t-1}, \mathbf{x}_0, \mathbf{h}_0)
      \;\|\;
      p_\theta(\mathbf{x}_{t-1} \mid  \mathbf{x}_t, \mathbf{h}_{t-1})
    \bigr)
  \Bigr]}_{\text{token KL (Pass 2)}}.
  \label{eq:kl_decomp}
\end{align}

The KL divergence splits into two terms \ref{eq:kl_decomp}, which we model using two stages in our architecture introduced in \ref{subsec:model}. 

\subsection{Model Architecture}
\label{subsec:model}
\subsubsection{Two Stage Model}
\captionsetup[figure]{skip=2pt}
\begin{figure}[h]
    \centering
    \includegraphics[width=0.9\textwidth]{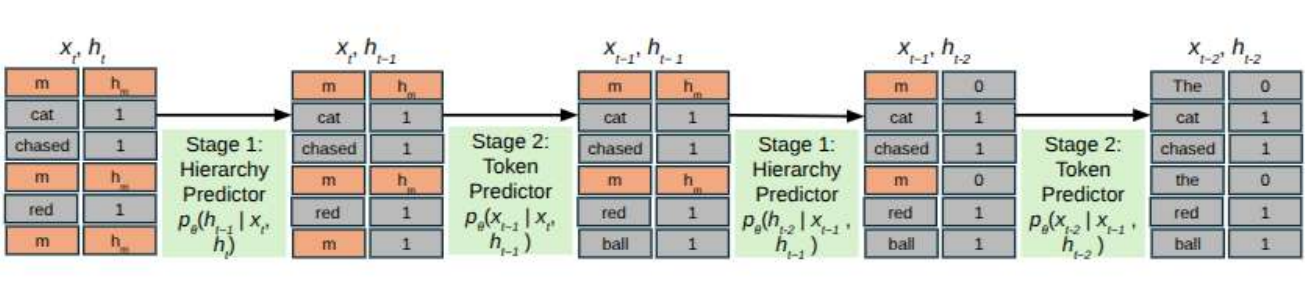}
    \caption{\textbf{HMDLM inference example.} Positions $i=0,3,5$ are masked in $(x_t,h_t)$. Step $t$: Stage 1 reveals $h_{t-1}^5=1$, then Stage 2 predicts $x_{t-1}^5=\text{ball}$. Step $t-1$: Stage 1 reveals $h_{t-2}^0=0$ and $h_{t-2}^3=0$, then Stage 2 predicts $x_{t-2}^0=\text{The}$ and $x_{t-2}^3=\text{the}$.}
    \label{fig:hmdlm_archiecture}
\end{figure}
To model the factorized posterior, we introduce the following two stage architecture \ref{fig:hmdlm_archiecture}. 

\textbf{First Stage:} Given the partially denoised sequence $(x_t, h_t)$ at time step $t$, the \textit{first stage} predicts the hierarchy distribution $g^i(t)$ at every masked position $i \in \mathcal{M}_t$, where $\mathcal{M}_t$ is the set of masked positions at time $t$; together with the noise schedule, this defines $p_\theta\!\left(h_{t-1}^i \mid x_t, h_t\right)$ in \eqref{eq:reverse_model}.
Masked positions are fed the masked state $\mathbf{s}_m$ as input to Stage I during both training and inference. During training, unmasked positions are provided with their gold pairs $(x_0^i, h_0^i)$ via teacher forcing. During inference, they are provided with the token and hierarchy sampled at the step when the position was first revealed, which remain fixed thereafter. 
At each inference step $t$, every masked position $i$ draws its next hierarchy state from \eqref{eq:reverse_model}: it is revealed at level $k \in \{0,\dots,K-1\}$ with probability $\lambda_k(t)\,g^i_k(t)$ and stays masked with probability $1-\pi_i(t)$. This single draw decides both whether position $i$ is revealed at this step and at which level, which forms the set of revealed positions $\mathcal{U}_t \subseteq \mathcal{M}_t$. 

\textbf{Second Stage:} The \textit{second stage} predicts the token distribution at each masked position $i$ at time step $t$: $p_\theta(x_{t-1}^i \mid x_t, h_{t-1})$. For masked positions, it receives as input $(x_t, h_{t-1})$ with the hierarchy at the reveal set filled in: positions $i \in \mathcal{U}_t$ are assigned gold hierarchy $h_0^i$, $(m,h_0^i)$ during training and the Stage-I sample $(m,h_{t-1}^i)$ during inference. For remaining masked positions, we pass the mask state $s_m$ = (m, $h_m$) as input. For unmasked positions, the token and hierarchy are copied directly from $x_t$ and $h_t$ without modification:
$p_\theta(s_{t-1}^i \mid s_t^i) = \mathrm{Cat}\left(s_{t-1}^i;\, s_t^i\right) \quad \text{if } s_t^i \neq \mathbf{s_m}$ during inference and passed $s^i_0$=($x^i_0$,$h^i_0$) during training. Both stages use weight tying, with separate output heads (hierarchy predictor head and token predictor head) attached to the final hidden layer and a softmax applied at the end of each head. This design reduces the parameter count while ensuring that both stages benefit from the same rich contextual representation. 

\textit{Note:} Across iterations, the token and hierarchy at each position, $s_t^i=(x_t^i,h_t^i)$, are masked and unmasked jointly through a shared Bernoulli variable \ref{eq:forward_bernoulli} . Within a single iteration, however, Stage~1 predicts the hierarchy before Stage~2 predicts the token. The partially unmasked state $(m,\,h_{t-1}^i)$ with $h_{t-1}^i\neq h_{\mathrm{m}}$ therefore arises only between the two stages of the same iteration.
\subsubsection{Per-level Noise Schedule and Hierarchy Embeddings}
\label{sec:noise_sch}

We use a different masking schedule for each hierarchy level $k$. Tokens that are more informative are given a higher hierarchy level and a slower masking schedule. Consequently, during the forward noising process, these tokens are masked more slowly than lower level tokens, preserving them for longer. They are denoised faster than other tokens during the reverse de-noising process. These tokens are unmasked earlier than lower-hierarchy tokens, allowing the model to condition on higher level tokens when predicting the remaining masked positions.

The hierarchy of a position is represented by a row of the learnable hierarchy embedding matrix $W \in \mathbb{R}^{(K+1) \times H}$, where $K$ is the number of hierarchy levels and
$H$ is the hidden dimension. Row $W_k$ is the embedding of level $k \in \{0,\dots,K-1\}$, and the last row $W_K$ is the embedding of the hierarchy mask state $h_{\mathrm{m}}$. The hierarchy embedding for each position $i$ is added to the token embedding at position $i$ and fed as input to the backbone.

\subsection{Training and Inference}
\label{sec:train}

\textbf{Training}
The HMDLM model is trained end-to-end across both the passes. We sample a time step $t$, and mask each token independently according to the token's hierarchy level masking schedule. The input to the Stage 1 is $(x_t, h_t)$ and it predicts the hierarchy probabilities - $p_\theta(h_{t-1}^i \mid x_t, h_t)$, we pass updated $(x_t, h_{t-1})$ to Stage 2 to get token distribution - $p_\theta(x_{t-1}^i \mid x_t, h_{t-1})$ for each masked position.  The final training loss is governed by: 
\begin{equation}
  \small
  \mathcal{L}
  = \mathbb{E}_{t,q}\!\left[
      \sum_{i=1}^{L}
      \mathbf{1}[x_t^i = m]\,
      \Bigl(
        \lambda'_{h_0^i}(t)\,\bigl(-\log f_{x_0^i}^i(t) - \beta_{\mathrm{h}} \log g_{h_0^i}^i(t)\bigr)
        + \beta_{\mathrm{e}}\bigl(
            {\textstyle\sum_{k} g^i_k(t)\,\lambda_k'(t)}
            - \lambda_{h_0^i}'(t)
          \bigr)
      \Bigr)
    \right]
  \label{eq:loss_total}
\end{equation}

where $\beta_h$ and $\beta_e$ are tunable hyperparameters,  $\lambda_k'(t) = \frac{-\alpha_k'(t)}{1-\alpha_k(t)}$, and $h_0^i$ is the gold hierarchy and $x_0^i$  is the gold token, $g^i_k(t)$ and $f^i_v(t)$ are the hierarchy-head and token-head probabilities of \eqref{eq:reverse_model} at position $i$. The full training procedure is given in Algorithm~\ref{alg:hmdlm_train}.

\textbf{Inference}
\label{sec:infer}
During the reverse process, we start from a fully masked sequence and iteratively unmask tokens at each time step $t$. At time step $t=T$ all the tokens are $[MASK]$ tokens and hence are assigned masked state, $s_m$. In Stage I, we sample hierarchy value for each masked token position $i$ at  time step $t$ from the predicted hierarchy probabilities as: 
\begin{equation}
 \resizebox{0.6\linewidth}{!}{$\displaystyle   h^i_{t-1} \sim \mathrm{Cat}\!\bigl(\lambda_0(t)\, g_0^i(t),\ \dots,\ \lambda_{K-1}(t)\, g_{K-1}^i(t),\ \underbrace{1-\pi_i(t)}_{\text{stay masked}}\bigr)$} 
\label{eq:pi}
\end{equation}
with $g^i(t)$ and $\pi_i(t)$ as in \eqref{eq:reverse_model}, and $\lambda_k(t)$ the per-step reveal probability of level $k$. Higher levels are masked more slowly, and for the schedules we use $\lambda_{k}(t) \le \lambda_{k'}(t)$ for $k < k'$ at every step at which level $k'$ is still masked (Appendix~\ref{app:ordering}). A position predicted to be at level $k'$ therefore receives a larger reveal mass $\lambda_{k'}(t)\,g^i_{k'}(t)$ and is revealed earlier. The full inference algorithm is given in the Appendix (Algorithm~\ref{alg:hmdlm_sample}).

%% file: experiments_orig.tex
\section{Experiments}
\label{sec:experiments}

In this section, we evaluate the proposed Hierarchical Masked Diffusion Language Model (HMDLM) along two axes:
(1) its ability to model language distributions, measured via perplexity, and
(2) its effectiveness as a puzzle solver, measured by accuracy. We consider three representative benchmarks: the One Billion Word Benchmark (LM1B)~\citep{chelba2013one} and WikiHow~\citep{koupaee2018wikihowlargescaletext} 
for generative modeling, and Sudoku~\citep{NEURIPS2018_7597}

\subsection{Language Generation Tasks}
\label{sec:common-setup}

\textbf{Setup.} For both LM1B and WikiHow, HMDLM adopts the MDLM backbone~\citep{sahoo2024simple}: $12$ transformer blocks, hidden size $768$, $12$ attention heads, rotary positional embeddings~\citep{su2024roformer}, and the dataset-dependent tokenizer for each corpus: BERT-base-uncased for LM1B and GPT-2 for WikiHow. The hierarchy-specific components ($K=2$ levels) are (i) a shared embedding matrix $W \in \mathbb{R}^{(K+1) \times H} = \mathbb{R}^{3 \times 768}$, with $K$ learnable level embeddings plus a learnable [MASK]-hierarchy row for unrevealed positions, and (ii) a hierarchy-prediction head that mirrors the token head's adaLN-modulated~\citep{peebles2023scalablediffusionmodelstransformers} final layer, mapping $\mathbb{R}^{768} \to \mathbb{R}^{2}$. Together these add $\approx 0.2$M parameters, about $0.15\%$ of the $139$M-parameter LM1B backbone and $0.12\%$ of the $170$M-parameter WikiHow backbone. The Sudoku architecture differs and is described in Section~\ref{sec:sudoku_section}. 

\textbf{Metrics.} We evaluate generation quality with four complementary measures. {Model PPL} assesses density modeling. {Generative PPL} assesses fluency of sampled text, but can be artificially lowered by leaning on a few high-likelihood phrases, producing repetitive, low-entropy output. We therefore read it alongside two diversity measures: {entropy}, where a value close to the reference or a high value indicates the model has not collapsed onto safe phrases, and MAUVE, which measures the similarity between the model-generated and reference text distributions. To corroborate the above metrics, we also run a pairwise LLM-as-judge study (GPT-5 judge) that
reports how often each model’s generations are preferred.

\subsubsection{LM1B Dataset}
\textbf{Task Description:} We evaluate the generative modeling capability of HMDLM by training it from scratch on LM1B, a standard dataset for language modeling. We compare test perplexity against several diffusion-based baselines: MDLM~\citep{sahoo2024simple}, SEDD~\citep{lou2024discrete}, the Anchored Diffusion Language Model (ADLM)~\citep{rout2025anchoreddiffusionlanguagemodel}, and DiffusionBERT~\citep{he2023diffusionbert}.

\setlength{\columnsep}{8pt}
\setlength{\intextsep}{5pt}

\begin{wraptable}{r}{0.45\textwidth}
  \vspace{-10pt}
  \centering
  \footnotesize
  \setlength{\tabcolsep}{3pt}
  \setlength{\abovecaptionskip}{2pt}
  \setlength{\belowcaptionskip}{2pt}
  \renewcommand{\arraystretch}{1.0}

  \caption{Test perplexities (PPL $\downarrow$) on LM1B. Best value is bolded.}
  \label{tab:ppl_33b}

  \resizebox{\linewidth}{!}{%
  \begin{tabular}{lc}
    \toprule
    {\scriptsize\textbf{Model}} & {\scriptsize\textbf{PPL ($\downarrow$)}} \\
    \midrule
    \multicolumn{2}{l}{\scriptsize\textit{33B tokens}} \\
    Diffusion BERT~\citep{he2023diffusionbert} & $\leq 63.78$ \\
    SEDD~\citep{lou2024discrete} & $\leq 32.79$ \\
    MDLM~\citep{sahoo2024simple} & $\leq 27.04$ \\
    ADLM~\citep{rout2025anchoreddiffusionlanguagemodel} & $\leq 26.40$ \\
    \textbf{HMDLM} & $\leq \mathbf{24.30}$ \\
    \midrule
    \multicolumn{2}{l}{\scriptsize\textit{65B tokens}} \\
    MDLM~\citep{sahoo2024simple} & $\leq 25.49$ \\
    ADLM~\citep{rout2025anchoreddiffusionlanguagemodel} & $\leq 24.46$ \\
    \textbf{HMDLM} & $\leq \mathbf{22.88}$ \\
   
    \bottomrule
  \end{tabular}}
  \vspace{-12pt}
\end{wraptable}
\textbf{Setup:} Using the shared language backbone (Section~\ref{sec:common-setup}), we set a context length of $128$ and train HMDLM from scratch for $1.25$M and $2.5$M steps ($\approx 33$B and $\approx 65$B tokens) on LM1B (see Section~\ref{sec:expected_mask_fraction}).

\textbf{Hierarchy Assignment:} We define two hierarchy levels on LM1B by in-sentence word frequency ($\tau$) following~\citet{rout2025anchoreddiffusionlanguagemodel}. Tokens appearing at most $\tau$ times in a sample are assigned to level $1$ (important words) and the rest to level $0$ (common words). Although ADLM reports numbers with $\tau=5$, this setting provides limited hierarchy separation: majority of the tokens get hierarchy 1 ($99.16\%$). Under $\tau=1$, around $79\%$ tokens get level 1, and the rest get assigned level 0 - yielding a more meaningful hierarchy. 

\setlength{\columnsep}{12pt}  
\setlength{\intextsep}{3pt}   

\begin{wraptable}{r}{0.55\textwidth}
  \vspace{-4pt}
  \centering
  \small
  \setlength{\tabcolsep}{3pt}
  \setlength{\abovecaptionskip}{2pt}
  \setlength{\belowcaptionskip}{2pt}

  \caption{LM1B generative evaluation on $65$B checkpoints. Mean $\pm$ std over 3 seeds.}
  \label{tab:lm_gen}

  \resizebox{\linewidth}{!}{%
  \begin{tabular}{lccc}
    \toprule
    Model & Gen.\ PPL ($\downarrow$) & MAUVE$^{\star}$ ($\uparrow$) & Entropy \\
    \midrule
    MDLM & $177.94{\scriptstyle\pm0.82}$
         & $0.66{\scriptstyle\pm0.03}$
         & $6.022{\scriptstyle\pm0.003}$ \\
    HMDLM & $\mathbf{83.35}{\scriptstyle\pm0.83}$
         & $\mathbf{0.76}{\scriptstyle\pm0.06}$
         & $5.876{\scriptstyle\pm0.014}$ \\
    \bottomrule
  \end{tabular}}
  \vspace{2pt}
  {\raggedright\scriptsize $^{\star}$MAUVE is reported here in $\%$.\par}
  \vspace{-4pt}
\end{wraptable}

\textbf{Noise Schedule:} We use a linear denoising schedule (identical across all models for a fair PPL comparison) with a \emph{level offset} $\delta = 0.20$, defined for HMDLM as
$\alpha_0(t) = 1-(1-\varepsilon)t$ and
$\alpha_1(t) = 1$ for $t\in[0,\delta]$, while
$\alpha_1(t) = 1-(1-\varepsilon)\frac{t-\delta}{1-\delta}$ for $t\in(\delta,1]$.

\textbf{Results:} We generate 1000 samples across three seeds for each model (128 token length and 256 steps). As can be seen from the Table \ref{tab:ppl_33b} and Table~\ref{tab:lm_gen}, HMDLM attains the best model PPL  and generative perplexity (measured using GPT-2 Large) among the diffusion baselines at
the $65$B-token budget, cutting generative PPL by
nearly $95$ points relative to MDLM \footnote{All the baseline results in Table \ref{tab:ppl_33b} are taken from~\citet{rout2025anchoreddiffusionlanguagemodel}. We tried training the ADLM model for comparison in Table \ref{tab:lm_gen}, but the gradients were exploding.}.
The reduction in generative PPL is accompanied by only a small decrease in entropy ($5.876$ vs. $6.022$), providing no clear evidence of severe diversity collapse. MAUVE scores, though low, are higher for HMDLM as compared to MDLM. An independent LLM-as-judge evaluation (prompt: \ref{sec:prompt}) is consistent with the results: in blinded pairwise comparisons with randomized A/B assignment ($100$ samples per seed, $3$ seeds, $2$ runs each; $600$ comparisons total),
the judge preferred HMDLM to MDLM $56\%$ to $44\%$, with no ties. Under the $\tau=5$ setting, HMDLM performs similar to MDLM, as nearly all real tokens are assigned to the same hierarchy level, leaving little scope for the hierarchical mechanism to provide an advantage.
\subsubsection{WikiHow Dataset}

\setlength{\columnsep}{12pt}   
\setlength{\intextsep}{0pt}    

\begin{wraptable}[8]{r}{0.6\textwidth}   
  \vspace{-4pt}
  \centering
  \small
  \setlength{\tabcolsep}{2pt}
  \setlength{\abovecaptionskip}{1pt}
  \setlength{\belowcaptionskip}{1pt}
  \caption{Model PPL ($\downarrow$) across sequence lengths on WikiHow. Best value is in bold.}
  \label{tab:model_ppl_lengths}

  \begin{tabular}{@{}lcccc@{}}
    \toprule
    $L$ & MDLM & HMDLM-P$^{\dagger}$ & ADLM & HMDLM-F$^{\ddagger}$ \\
    \midrule
    $256$  & $24.60\pm0.22$ & $23.89\pm0.11$ & $22.05\pm0.21$ & $\mathbf{17.62\pm0.12}$ \\
    $512$  & $21.04\pm0.06$ & $21.76\pm0.15$ & $20.15\pm0.10$ & $\mathbf{16.64\pm0.06}$ \\
    $1024$ & $22.65\pm0.04$ & $22.58\pm0.04$ & $20.92\pm0.05$ & $\mathbf{16.90\pm0.09}$ \\
    \bottomrule
  \end{tabular}
  \par\vspace{1pt}
  {\raggedright\scriptsize
   $^{\dagger}$HMDLM-P: position-based hierarchy. \;
   $^{\ddagger}$HMDLM-F: frequency-based hierarchy.\par}
  \vspace{-12pt}   
\end{wraptable}
\textbf{Task Description:} We next evaluate generative modeling on WikiHow. We initialize HMDLM from the publicly released MDLM checkpoint pretrained on
OpenWebText (OWT)~\citep{Gokaslan2019OpenWeb}, which we refer to as MDLM-OWT,
and fully fine-tune it on WikiHow.
As baselines, we fine-tune the MDLM-OWT and ADLM-OWT checkpoints on WikiHow under an identical setting.


\textbf{Setup:} We reuse the shared language backbone (Section~\ref{sec:common-setup}), fully fine-tuning from the MDLM-OWT checkpoint. We experiment with maximum sequence lengths of $L \in \{256,512,1024\}$, filtering WikiHow to corresponding token length. The train, val, and test set corresponding to each token length $L$ is kept consistent across all models for fair comparison. During inference, we use $256$ sampling steps for all three sequence lengths. All the baseline models are trained for $30$ epochs. Both the HMDLM variants are trained slightly longer because of lower expected masking fraction. Detailed discussion in Appendix \ref{app:exp_masking_frac_wikihow}.

\setlength{\columnsep}{12pt}  
\setlength{\intextsep}{2pt}   
\begin{wraptable}{r}{0.6\textwidth}
  \vspace{-2pt}
  \centering
  \setlength{\tabcolsep}{4pt}
  \setlength{\abovecaptionskip}{1pt}
  \setlength{\belowcaptionskip}{1pt}
  \caption{Generative evaluation on WikiHow across sequence lengths $L \in \{256, 512, 1024\}$. HMDLM-P/F: position-/frequency-based hierarchy. Gold is the reference data.}
  \label{tab:gen_eval_lengths}
  \resizebox{\linewidth}{!}{%
  \begin{tabular}{@{}l ccc ccc ccc@{}}
    \toprule
    & \multicolumn{3}{c}{Gen.\ PPL}
    & \multicolumn{3}{c}{MAUVE ($\uparrow$)}
    & \multicolumn{3}{c}{Entropy} \\
    \cmidrule(lr){2-4}\cmidrule(lr){5-7}\cmidrule(lr){8-10}
    Model & 256 & 512 & 1024 & 256 & 512 & 1024 & 256 & 512 & 1024 \\
    \midrule
    \textit{Gold} & 25.2 & 21.0 & 20.8 & 1.00 & 1.00 & 1.00 & 5.86 & 6.42 & 6.70 \\
    \midrule
    MDLM    & 76.0 & 67.4 & 62.7 & 0.33 & 0.11 & 0.12 & 6.47 & 7.10 & 7.60 \\
    ADLM    & 84.0 & 82.1 & 93.81   & 0.40 & 0.21 & 0.08   & 6.49 & 7.17 & 7.77   \\
    HMDLM-P & 23.1 & 17.6 & 14.6 & 0.38 & 0.13 & 0.07 & 5.91 & 6.26 & 6.45 \\
    HMDLM-F & 29.2 & 25.4 & $19.92$   & \textbf{0.52} & \textbf{0.41} & \textbf{0.20} & 5.98 & 6.60 & $6.83$ \\
    \bottomrule
  \end{tabular}}
\end{wraptable}
\textbf{Hierarchy Assignment:} We consider two hierarchy-assignment strategies. In the \emph{position-based} scheme, we assign level $1$ to summary tokens and level $0$ to the rest; here we compare directly against the fine-tuned MDLM model, since ADLM defines its anchor tokens by frequency rather than position and is therefore not comparable under this assignment. In the \emph{Frequency-based} scheme, we reuse the frequency rule from LM1B, assigning level $1$ to tokens occurring at most once in a sentence and level $0$ otherwise; this is directly comparable with ADLM, as we apply the same rule to define its anchor positions.

\textbf{Noise Schedule:} We use the power-function schedule $\alpha_h(t) = 1 - (1-\epsilon)\, t^{\omega_h}$ with $\omega_h[0] = 1.0$ and $\omega_h[1] = 2.0$. Setting $\omega_h[0] = 1.0$ recovers the log-linear schedule of MDLM.

\textbf{Results:} HMDLM-F (Frequency based hierarchy) achieves the best model PPL and its generative PPL is closest to reference distribution. HMDLM-P (Position based hierarchy) achieves the best gen PPL, on all three generation lengths (Tables~\ref{tab:model_ppl_lengths} and \ref{tab:gen_eval_lengths}), cutting gen PPL by nearly $50$ points relative to the baselines. The improvement is not repetition-driven: MAUVE is higher for HMDLM-F at every length, and entropy stays close to that of the reference dataset, indicating a closer match in terms of token distribution. LLM-as-judge evaluation (prompt: \ref{sec:prompt}) is consistent with the results: in blinded comparisons (400
samples, 2 runs each; 800 comparisons total), the judge preferred HMDLM-P $40.75\%$ times, HMDLM-F $31.88\%$ times, ADLM $15.25\%$, and MDLM $12.12\%$. Therefore, HMDLM combined was preferred more than twice the times baselines were selected. We also find that, for this dataset, HMDLM-F has lower model PPL on test set, better MAUVE scores and entropy closer to that of reference data than the HMDLM-P for all lengths. Refer to Section \ref{sec:wikihow_qual} in the Appendix for a qualitative analysis

\textbf{Position- vs. frequency-based hierarchy:} On this dataset, HMDLM-F outperforms HMDLM-P in model PPL, MAUVE, and entropy (closer to reference distribution) at all lengths. We hypothesize that generating key tokens first matters more than their positions, since these tokens provide enough context to decode the rest of the text. This is domain-dependent: for tasks like Sudoku, position-based hierarchy is likely essential. Both variants improve over MDLM, and ADLM, which is implicitly frequency-driven, is consistently outperformed by HMDLM-F. This highlights the benefit of our principled treatment of hierarchy in diffusion models.
\subsection{Sudoku Dataset}
\label{sec:sudoku_section}

\textbf{Task Description:} Sudoku is a useful test case for structured reasoning. Some cells can be filled directly from the given numbers, while others can be resolved only after several other cells have been filled; revealing the easy cells first therefore makes the harder cells easier to predict. In contrast, the current diffusion models (MDLM and MGDM~\citep{ye2025beyond}) reveal cells without exploiting such an order. Each puzzle has $81$ cells and some given cells. We use the Sudoku dataset of~\citet{NEURIPS2018_7597} 
with $180$k training, $18$k validation, and $18$k test puzzles. The given cells are never masked and are excluded from the loss.
\setlength{\columnsep}{7pt}   
\setlength{\intextsep}{4pt}   
\begin{wraptable}[11]{r}{0.42\textwidth}

  \centering
  \footnotesize
  \setlength{\tabcolsep}{3pt}
  \setlength{\abovecaptionskip}{2pt}
  \setlength{\belowcaptionskip}{2pt}
  \caption{Sudoku-hard results: 25M parameter models under stochastic \& greedy decoding.}
  \label{tab:sudoku_24m}
  \begin{tabular}{llcc}
    \toprule
    Metric & Model & stochastic & Greedy \\
    \midrule
    \multirow{2}{*}{Board Acc.}
      & MDLM  & $0.81$ & $0.83$ \\
      & HMDLM & $\mathbf{0.86}$ & $\mathbf{0.88}$ \\
    \midrule
    \multirow{2}{*}{Cell Acc.}
      & MDLM  & $0.94$ & $0.94$ \\
      & HMDLM & $\mathbf{0.96}$ & $\mathbf{0.97}$ \\
    \bottomrule
  \end{tabular}
\end{wraptable}

\textbf{Hierarchy labels.}
Starting from the given cells, we repeatedly fill the masked cells that can be determined directly. A cell is assigned a level according to the round in which it is first determined. We use six levels ($K=6$): level $5$ contains cells determined directly from the given cells, while lower levels contain cells that require more rounds of solving. Level $0$ contains cells that are determined only after six or more rounds, or that are never determined by these rules and therefore require search. At test time, the levels of blank cells are predicted by the hierarchy head. Architecture details for both variants (25M and 6M) are provided in Appendix~\ref{sec:sudoku_qual}.

\setlength{\columnsep}{5pt}   
\setlength{\intextsep}{4pt}   
\begin{wraptable}[10]{r}{0.48\columnwidth}
  \vspace{4pt}
  \centering
  \small
  \setlength{\tabcolsep}{3pt}
  \setlength{\abovecaptionskip}{2pt}
  \setlength{\belowcaptionskip}{2pt}
  \caption{Sudoku-hard test set ($18$k puzzles). Results are from $6$M-parameter models trained for $100$k steps with $T=81$ and confidence-based sampling. Mean $\pm$ std over $3$ sampling seeds.}
  \label{tab:sudoku_static_hierarchy}
  \begin{tabular}{lcc}
    \toprule
    Model & Board Acc. & Cell Acc. \\
    \midrule
    MGDM & $63.07 \pm 0.18$ & $85.79 \pm 0.10$ \\ 
    HMDLM & $\mathbf{71.69 \pm 0.26}$ & $\mathbf{91.13 \pm 0.07}$ \\
    \bottomrule
  \end{tabular}
\end{wraptable}
\textbf{Results.}
HMDLM consistently outperforms the baseline in both solve rate and cell accuracy: for the $25$M model it wins under both stochastic and greedy decoding (Table~\ref{tab:sudoku_24m}), and the same advantage holds for the compact $6$M model (Table~\ref{tab:sudoku_static_hierarchy}). We also analyze the cell accuracy for each hierarchy level (Table~\ref{tab:sudoku-cell-levels}), HMDLM shows the largest gains, up to $8\%$ over MDLM for level 0 cells (or the cells that are most difficult to fill). Additionally, we ran a qualitative analysis on Sudoku puzzles and found that, for the $25$M model, among 287 puzzles solved by HMDLM but not MDLM across all three seeds, $45.6\%$ showed MDLM making an early commitment to a search cell, while HMDLM delayed filling the same cell until fewer candidates remained. Details in Appendix \ref{sec:sudoku_qual}

%% file: conclusion.tex
\section{Conclusion and Limitations}
\label{sec:conclusion}

We introduced HMDLM, a hierarchical extension of masked diffusion language models that assigns priority to token positions and incorporates it into both the forward and reverse processes. By using level-specific noise schedules and a two-stage architecture that first predicts hierarchy and then tokens, HMDLM enables structured and controllable generation while remaining lightweight and modular. Our model consistently outperforms strong diffusion baselines by decoding semantically important tokens early and using them to better condition the rest of the sequence. While this improvement comes at the cost of additional computation due to the two-stage design, our results highlight hierarchy-aware generation as a promising direction for diffusion-based language models.


%% file: appendix.tex
\section{Discrete-Time ELBO}
\label{app:discrete_elbo}
\subsection{Joint Generative Model $p_\theta(\mathbf{x}_{0}, \mathbf{h}_{0})$}
\label{app:joint_model}

\paragraph{Hierarchy is jointly modelled with tokens.}
We represent the state of the model at time step $t$ as $\mathbf{s_t}=(\mathbf{x_t}, \mathbf{h_t})$. When token $i$
is masked ($x_t^i = M$), \emph{both} the token and its hierarchy level are masked: the
token is replaced by \texttt{[MASK]} and $h_t^i$ is set to masked hierarchy
state ($h_{MASK}$). \hmdlm{} therefore targets
the {joint} distribution:
\begin{equation}
  \log p_\theta(\mathbf{x}_{0}, \mathbf{h}_{0}).
  \label{eq:joint_target}
\end{equation}

For $s < t$ and a masked position ($x_t^i = m$), both $x_0^i$ and $h_0^i$ are revealed
together upon unmasking, with the hierarchy fixed consistently to $h_0^i = f(x_0^i)$: 
\begin{equation}
  (x_s^i,\; h_s^i) =
  \begin{cases}
    (x_0^i,\; h_0^i)
      & \text{with prob.}\quad
        \dfrac{\alpha_{h_0^i}(s) - \alpha_{h_0^i}(t)}{1 - \alpha_{h_0^i}(t)}
        \quad\text{(jointly revealed)} \\[12pt]
    (m,\; h_{mask})
      & \text{with prob.}\quad
        \dfrac{1 - \alpha_{h_0^i}(s)}{1 - \alpha_{h_0^i}(t)}
        \quad\text{(jointly remain masked)}
  \end{cases}
  \label{eq:joint_posterior}
\end{equation}

\subsection{Step 1 --- Introduce the Joint Diffusion Process}

We target the \emph{joint} distribution
$\log p_\theta(\mathbf{x}_{0}, \mathbf{h}_{0})$. The joint
$(\mathbf{x}_t, \mathbf{h}_t)$ is the diffusion state at time $t$: when $x_t^i = M$, both
the token and its hierarchy label are masked and unobserved; when $x_t^i \neq M$, both are
revealed together. Introduce the latent noisy sequence of joint states
$(\mathbf{x}_{1:T}, \mathbf{h}_{1:T})$ at $T$ discrete time steps $t_1 < \cdots < t_T = 1$:
\begin{equation}
  \log p_\theta(\mathbf{x}_{0}, \mathbf{h}_{0})
  = \log \int p_\theta(\mathbf{x}_{0:T}, \mathbf{h}_{0:T})\; d\mathbf{x}_{1:T}\, d\mathbf{h}_{1:T}.
  \label{eq:log_p}
\end{equation}

Because each denoising step depends only on the current state and not on the whole history, the joint model breaks into a product of one-step reverse predictions:
\begin{equation}
p_\theta(\mathbf{x}_{0:T}, \mathbf{h}_{0:T}) = p(\mathbf{x}_T, \mathbf{h}_T)\prod_{t=1}^{T} p_\theta(\mathbf{x}_{t-1}, \mathbf{h}_{t-1}\mid \mathbf{x}_t, \mathbf{h}_t)\end{equation}
We now split each reverse step into two parts: first predict the hierarchy, then predict the token given that hierarchy. We also drop $\mathbf{h}_t$ from the token prediction, since $\mathbf{h}_{t-1}$ already carries the hierarchy information needed. Finally, since the last state is fully masked and fixed, its prior splits into a product $p(\mathbf{x}_T)\,p(\mathbf{h}_T)$.
This gives:

\begin{equation}
  p_\theta(\mathbf{x}_{0:T}, \mathbf{h}_{0:T})
  = p(\mathbf{x}_T)\, p(\mathbf{h}_T)
    \prod_{t=1}^{T}
    \underbrace{p_\theta(\mathbf{h}_{t-1} \mid \mathbf{h}_t,\, \mathbf{x}_t)}_{\text{Pass 1: hierarchy}}
    \cdot
    \underbrace{p_\theta(\mathbf{x}_{t-1} \mid \mathbf{h}_{t-1},\, \mathbf{x}_t)}_{\text{Pass 2: tokens}}.
  \label{eq:joint_gen}
\end{equation}
Multiply and divide by the forward process
$q(\mathbf{x}_{1:T}, \mathbf{h}_{1:T} \mid \mathbf{x}_0, \mathbf{h}_0)$ in \ref{eq:log_p}:

\begin{equation}
  \log p_\theta(\mathbf{x}_{0}, \mathbf{h}_{0})
  = \log \mathbb{E}_{q(\mathbf{x}_{1:T}, \mathbf{h}_{1:T} \mid \mathbf{x}_0, \mathbf{h}_0)}
    \!\left[
      \frac{p_\theta(\mathbf{x}_{0:T}, \mathbf{h}_{0:T})}
           {q(\mathbf{x}_{1:T}, \mathbf{h}_{1:T} \mid \mathbf{x}_0, \mathbf{h}_0)}
    \right].
\end{equation}

\subsection{Step 2 --- Apply Jensen's Inequality}

Since $\log$ is concave, Jensen's inequality gives:
\begin{equation}
  \log p_\theta(\mathbf{x}_{0}, \mathbf{h}_{0})
  \;\geq\;
  \mathbb{E}_{q(\mathbf{x}_{1:T}, \mathbf{h}_{1:T} \mid \mathbf{x}_0, \mathbf{h}_0)}
  \!\left[
    \log \frac{p_\theta(\mathbf{x}_{0:T}, \mathbf{h}_{0:T})}
              {q(\mathbf{x}_{1:T}, \mathbf{h}_{1:T} \mid \mathbf{x}_0, \mathbf{h}_0)}
  \right].
  \label{eq:jensen}
\end{equation}

\subsection{Step 3 --- Expand the Log of the Joint Model}

The forward process factorises as:
\begin{equation}
  q(\mathbf{x}_{1:T}, \mathbf{h}_{1:T} \mid \mathbf{x}_0, \mathbf{h}_0)
  = \prod_{t=1}^{T} q(\mathbf{x}_t, \mathbf{h}_t \mid \mathbf{x}_{t-1}, \mathbf{h}_{t-1}).
\end{equation}
Taking the log of~\eqref{eq:joint_gen} gives:
\begin{align}
  \log p_\theta(\mathbf{x}_{0:T}, \mathbf{h}_{0:T})
  = \log p(\mathbf{x}_T) + \log p(\mathbf{h}_T)
    &+ \underbrace{\sum_{t=1}^{T} \log p_\theta(\mathbf{h}_{t-1} \mid \mathbf{h}_t,\, \mathbf{x}_t)}_{\text{Pass 1: hierarchy}}
  \nonumber\\
    &+ \underbrace{\sum_{t=1}^{T} \log p_\theta(\mathbf{x}_{t-1} \mid \mathbf{h}_{t-1},\,  \mathbf{x}_t)}_{\text{Pass 2: tokens}}.
  \label{eq:log_joint_gen}
\end{align}
Substituting into~\eqref{eq:jensen}:
\begin{align}
  \log p_\theta(\mathbf{x}_0, \mathbf{h}_0)
  \;\geq\;
  \mathbb{E}_q\!\Bigl[
    &\log p(\mathbf{x}_T) + \log p(\mathbf{h}_T)
  \nonumber\\
    &+ \sum_{t=1}^{T} \log p_\theta(\mathbf{h}_{t-1} \mid \mathbf{h}_t,\, \mathbf{x}_t)
  \nonumber\\
    &+ \sum_{t=1}^{T} \log p_\theta(\mathbf{x}_{t-1} \mid \mathbf{h}_{t-1},\,  \mathbf{x}_t)
  \nonumber\\
    &- \sum_{t=1}^{T} \log q(\mathbf{x}_t, \mathbf{h}_t \mid \mathbf{x}_{t-1}, \mathbf{h}_{t-1})
  \Bigr].
  \label{eq:elbo_split}
\end{align}

\subsection{Step 4 --- Telescope the Forward Terms}

\paragraph{Key identity.}
By Bayes' rule on the joint forward process:
\begin{equation}
  q(\mathbf{x}_{t-1}, \mathbf{h}_{t-1} \mid \mathbf{x}_t, \mathbf{h}_t, \mathbf{x}_0, \mathbf{h}_0)
  = \frac{q(\mathbf{x}_t, \mathbf{h}_t \mid \mathbf{x}_{t-1}, \mathbf{h}_{t-1})\;
          q(\mathbf{x}_{t-1}, \mathbf{h}_{t-1} \mid \mathbf{x}_0, \mathbf{h}_0)}
         {q(\mathbf{x}_t, \mathbf{h}_t \mid \mathbf{x}_0, \mathbf{h}_0)}.
  \label{eq:bayes_id}
\end{equation}
Rearranging:
\begin{equation}
\begin{split}
  \log q(\mathbf{x}_t, \mathbf{h}_t \mid \mathbf{x}_{t-1}, \mathbf{h}_{t-1})
  &= \log q(\mathbf{x}_{t-1}, \mathbf{h}_{t-1} \mid \mathbf{x}_t, \mathbf{h}_t, \mathbf{x}_0, \mathbf{h}_0) \\
  &\quad + \log q(\mathbf{x}_t, \mathbf{h}_t \mid \mathbf{x}_0, \mathbf{h}_0)
  - \log q(\mathbf{x}_{t-1}, \mathbf{h}_{t-1} \mid \mathbf{x}_0, \mathbf{h}_0).
\end{split}
\label{eq:log_bayes}
\end{equation}

\paragraph{Telescoping.}
Substituting~\eqref{eq:log_bayes} into
$\sum_{t=1}^{T} \log q(\mathbf{x}_t, \mathbf{h}_t \mid \mathbf{x}_{t-1}, \mathbf{h}_{t-1})$
and using $q(\mathbf{x}_0, \mathbf{h}_0 \mid \mathbf{x}_0, \mathbf{h}_0) = 1$:
\begin{equation}
  \sum_{t=1}^{T} \log q(\mathbf{x}_t, \mathbf{h}_t \mid \mathbf{x}_{t-1}, \mathbf{h}_{t-1})
  = \sum_{t=1}^{T} \log q(\mathbf{x}_{t-1}, \mathbf{h}_{t-1} \mid \mathbf{x}_t, \mathbf{h}_t, \mathbf{x}_0, \mathbf{h}_0)
  + \log q(\mathbf{x}_T, \mathbf{h}_T \mid \mathbf{x}_0, \mathbf{h}_0).
  \label{eq:telescope}
\end{equation}

\paragraph{Substitute back.}
Applying~\eqref{eq:telescope} to~\eqref{eq:elbo_split}, grouping $t=1$ separately,
and identifying each $(\log p_\theta - \log q)$ pair as $-D_{\mathrm{KL}}(q\|p)$:
\begin{align}
  \log p_\theta(\mathbf{x}_{0}, \mathbf{h}_{0})
  \;\geq\;
  &\underbrace{\mathbb{E}_q\!\bigl[
      \log p_\theta(\mathbf{x}_{0}, \mathbf{h}_{0} \mid \mathbf{x}_1, \mathbf{h}_1)
    \bigr]}_{L_0}
  \nonumber\\[4pt]
  &- \sum_{t=2}^{T}
    \underbrace{\mathbb{E}_q\!\Bigl[
      D_{\mathrm{KL}}\bigl(
        q(\mathbf{x}_{t-1}, \mathbf{h}_{t-1} \mid \mathbf{x}_t, \mathbf{h}_t, \mathbf{x}_0, \mathbf{h}_0)
        \;\|\;
        p_\theta(\mathbf{x}_{t-1}, \mathbf{h}_{t-1} \mid \mathbf{x}_t, \mathbf{h}_t)
      \bigr)
    \Bigr]}_{L_t}
  \nonumber\\[4pt]
  &- \underbrace{D_{\mathrm{KL}}\bigl(
      q(\mathbf{x}_T, \mathbf{h}_T \mid \mathbf{x}_0, \mathbf{h}_0)
      \;\|\;
      p(\mathbf{x}_T, \mathbf{h}_T)
    \bigr)}_{L_T}.
  \label{eq:discrete_elbo}
\end{align}
Here $p(\mathbf{x}_T, \mathbf{h}_T) = p(\mathbf{x}_T)\,p(\mathbf{h}_T)$ since the terminal
prior factorises. The reverse model factorises as
\begin{equation}
  p_\theta(\mathbf{x}_{t-1}, \mathbf{h}_{t-1} \mid \mathbf{x}_t, \mathbf{h}_t)
  = p_\theta(\mathbf{h}_{t-1} \mid \mathbf{x}_t, \mathbf{h}_t)\cdot
    p_\theta(\mathbf{x}_{t-1} \mid \mathbf{x}_t, \mathbf{h}_{t-1})
  \label{eq:reverse_factorization}
\end{equation}

We approximate the reverse process $p_\theta(\mathbf{x}_{t-1},\mathbf{h}_{t-1} | \mathbf{x}_t, \mathbf{h}_t)$
using the neural network $s_\theta(\mathbf{x}_t, \mathbf{h}_t, t)$, which is composed of two joined networks - $p_\theta(\mathbf{h}_{t-1} \mid \mathbf{x}_t, \mathbf{h}_t)$ is modeled as $h_\theta(\mathbf{x}_t, \mathbf{h}_t, t)$ and $p_\theta(\mathbf{x}_{t-1} \mid \mathbf{x}_t, \mathbf{h}_{t-1})$ is modeled as $x_\theta(\mathbf{x}_t, \mathbf{h}_{t-1}, t)$.

\subsection{Step 5 --- Decompose the Joint KL Term $L_t$ (Full Derivation)}

Each $L_t$ is a KL over the joint state $(\mathbf{x}_{t-1}, \mathbf{h}_{t-1})$,
so that the term of interest is
$L_t = D_{\mathrm{KL}}\bigl(q(\mathbf{x}_{t-1}, \mathbf{h}_{t-1} \mid \mathbf{x}_t,\mathbf{h}_t,\mathbf{x}_0,\mathbf{h}_0)
  \,\|\, p_\theta(\mathbf{x}_{t-1}, \mathbf{h}_{t-1} \mid \mathbf{x}_t, \mathbf{h}_t)\bigr)$.

\paragraph{Chain rule of KL divergence.}
For any two joint distributions $q(a,b\mid c)$ and $p(a,b\mid d)$ that are each
factorised in the \emph{same} variable order, the KL divergence obeys the exact
identity
\begin{equation}
  D_{\mathrm{KL}}\bigl(q(a,b\mid c)\,\|\,p(a,b\mid d)\bigr)
  =
  D_{\mathrm{KL}}\bigl(q(a\mid c)\,\|\,p(a\mid d)\bigr)
  +
  \mathbb{E}_{q(a\mid c)}\!\Bigl[
    D_{\mathrm{KL}}\bigl(q(b\mid a,c)\,\|\,p(b\mid a,d)\bigr)
  \Bigr].
  \label{eq:kl_chain_rule}
\end{equation}

\paragraph{Hierarchy-first factorisation.}
We apply~\eqref{eq:kl_chain_rule} with $a = \mathbf{h}_{t-1}$ (hierarchy) and
$b = \mathbf{x}_{t-1}$ (token), matching the two-stage structure of the model.
The true posterior factorises by the product rule ,
\begin{equation}
  q(\mathbf{x}_{t-1}, \mathbf{h}_{t-1} \mid \mathbf{x}_t,\mathbf{h}_t,\mathbf{x}_0,\mathbf{h}_0)
  = q(\mathbf{h}_{t-1} \mid \mathbf{x}_t,\mathbf{h}_t,\mathbf{x}_0,\mathbf{h}_0)\,
    q(\mathbf{x}_{t-1} \mid \mathbf{x}_t,\mathbf{h}_{t-1} ,\mathbf{x}_0,\mathbf{h}_0), 
  \label{eq:q_factor}
\end{equation}
Note: We drop $h_t$ since $h_{t-1}$ contains the latest state of the model.

The reverse model is \emph{defined} to
factorise in the same order - Pass~1 predicts $\mathbf{h}_{t-1}$ and Pass~2
predicts $\mathbf{x}_{t-1}$ conditioned on it 
\begin{equation}
  p_\theta(\mathbf{x}_{t-1}, \mathbf{h}_{t-1} \mid \mathbf{x}_t,\mathbf{h}_t)
  = \underbrace{p_\theta(\mathbf{h}_{t-1} \mid \mathbf{x}_t,\mathbf{h}_t)}_{\text{Pass 1: hierarchy head}}\,
    \underbrace{p_\theta(\mathbf{x}_{t-1} \mid \mathbf{x}_t,\mathbf{h}_{t-1})}_{\text{Pass 2: token head}}.
  \label{eq:p_factor}
\end{equation}

\paragraph{Apply the chain rule.}
Substituting the factorisations~\eqref{eq:q_factor} and~\eqref{eq:p_factor} into
the identity~\eqref{eq:kl_chain_rule} yields the two-term decomposition
\begin{align}
  &D_{\mathrm{KL}}\bigl(
    q(\mathbf{x}_{t-1}, \mathbf{h}_{t-1} \mid \mathbf{x}_t,\mathbf{h}_t,\mathbf{x}_0,\mathbf{h}_0)
    \;\|\;
    p_\theta(\mathbf{x}_{t-1}, \mathbf{h}_{t-1} \mid \mathbf{x}_t,\mathbf{h}_t)
  \bigr)
  \nonumber\\[4pt]
  &\quad=
  \underbrace{D_{\mathrm{KL}}\bigl(
    q(\mathbf{h}_{t-1} \mid \mathbf{x}_t,\mathbf{h}_t,\mathbf{x}_0,\mathbf{h}_0)
    \;\|\;
    p_\theta(\mathbf{h}_{t-1} \mid \mathbf{x}_t,\mathbf{h}_t)
  \bigr)}_{\text{hierarchy KL (Pass 1)}}
  \nonumber\\[4pt]
  &\qquad+
  \underbrace{\mathbb{E}_{q(\mathbf{h}_{t-1} \mid \mathbf{x}_t,\mathbf{h}_t,\mathbf{x}_0,\mathbf{h}_0)}\!\Bigl[
    D_{\mathrm{KL}}\bigl(
      q(\mathbf{x}_{t-1} \mid \mathbf{x}_t,\mathbf{h}_{t-1},\mathbf{x}_0,\mathbf{h}_0)
      \;\|\;
      p_\theta(\mathbf{x}_{t-1} \mid  \mathbf{x}_t,\mathbf{h}_{t-1})
    \bigr)
  \Bigr]}_{\text{token KL (Pass 2)}}.
  \label{eq:kl_decomp_app}
\end{align}

\subsection{Step 6 --- Simplify the Terminal Term $L_T$}

At $t = T$, every token and hierarchy label is fully masked with probability
$1 - \alpha_k(1) \approx 1$. The joint prior $p(\mathbf{x}_T, \mathbf{h}_T)$ is therefore
also the fully-masked state, so
\begin{equation}
  L_T = D_{\mathrm{KL}}\bigl(
      q(\mathbf{x}_T, \mathbf{h}_T \mid \mathbf{x}_0, \mathbf{h}_0)
      \;\|\;
      p(\mathbf{x}_T, \mathbf{h}_T)
    \bigr)
  \approx 0.
\end{equation}
We drop this term.

\subsection{Step 7 --- Full Joint Discrete ELBO}

Dropping $L_T = 0$:
\begin{align}
\log p_\theta(\mathbf{x}_{0}, \mathbf{h}_{0})
\;\geq\;
&\underbrace{\mathbb{E}_q\!\bigl[
    \log p_\theta(\mathbf{x}_{0}, \mathbf{h}_{0} \mid \mathbf{x}_1, \mathbf{h}_1)
  \bigr]}_{L_0}
  \nonumber\\[4pt]
&- \sum_{t=2}^{T}
  \underbrace{\mathbb{E}_q\!\Bigl[
    D_{\mathrm{KL}}\bigl(
      q(\mathbf{h}_{t-1} \mid \mathbf{x}_t, \mathbf{h}_t, \mathbf{x}_0, \mathbf{h}_0)
      \;\|\;
      p_\theta(\mathbf{h}_{t-1} \mid \mathbf{x}_t, \mathbf{h}_t)
    \bigr)
  \Bigr]}_{L_t^{\text{hier}}\ \text{(hierarchy head)}}
  \nonumber\\[4pt]
&- \sum_{t=2}^{T}
  \underbrace{\mathbb{E}_q\!\biggl[
    \mathbb{E}_{q(\mathbf{h}_{t-1} \mid \mathbf{x}_t, \mathbf{h}_t, \mathbf{x}_0, \mathbf{h}_0)}\!\Bigl[
      D_{\mathrm{KL}}\bigl(
        q(\mathbf{x}_{t-1} \mid \mathbf{x}_t, \mathbf{h}_{t-1},   \mathbf{x}_0, \mathbf{h}_0)
        \;\|\;
        p_\theta(\mathbf{x}_{t-1} \mid \mathbf{x}_t, \mathbf{h}_{t-1})
      \bigr)
    \Bigr]
  \biggr]}_{L_t^{\text{tok}}\ \text{(token head)}}.
  \label{eq:elbo_final}
\end{align}

\subsection{Step 8 --- Expand the Hierarchy KL (Pass 1)}

From~\eqref{eq:elbo_final}, the hierarchy contribution at reverse step $t$ is:
\begin{equation}
L_t^{\mathrm{hier}}
=
\mathbb{E}_q\!\left[
D_{\mathrm{KL}}\!\left(
q(\mathbf{h}_{t-1}\mid
\mathbf{x}_t,\mathbf{h}_t,\mathbf{x}_0,\mathbf{h}_0)
\;\|\;
p_\theta(\mathbf{h}_{t-1}\mid
\mathbf{x}_t,\mathbf{h}_t)
\right)
\right].
\label{eq:Lt_hier_start}
\end{equation}

Since the forward process factorizes independently across sequence positions, and
the Pass-1 hierarchy head predicts each position independently given the noisy
state $(\mathbf{x}_t,\mathbf{h}_t)$,
\begin{align}
q(\mathbf{h}_{t-1}\mid\cdots)
&=
\prod_{i=1}^{L}
q(h_{t-1}^i\mid\cdots),\\
p_\theta(\mathbf{h}_{t-1}\mid\mathbf{x}_t,\mathbf{h}_t)
&=
\prod_{i=1}^{L}
p_\theta(h_{t-1}^i\mid\mathbf{x}_t,\mathbf{h}_t).
\end{align}
The joint KL therefore decomposes into a sum of independent per-position KLs,
\begin{equation}
L_t^{\mathrm{hier}}
=
\mathbb{E}_q\!\left[
\sum_{i=1}^{L}
D_{\mathrm{KL}}
\!\left(
q(h_{t-1}^i\mid\cdots)
\;\|\;
p_\theta(h_{t-1}^i\mid
\mathbf{x}_t,\mathbf{h}_t)
\right)
\right].
\label{eq:Lt_hier_positionwise}
\end{equation}

\paragraph{Notation.}
Let
\begin{equation}
g_h^i(t)
:=
[h_\theta^i(\mathbf{x}_t,\mathbf{h}_t,t)]_h
,
\qquad
h\in\{1,\ldots,K\},
\qquad
\sum_{h=1}^{K}g_h^i(t)=1,
\label{eq:g_def}
\end{equation}
denote the Pass-1 softmax distribution over the $K$ hierarchy classes at position
$i$. For convenience we write
\begin{equation}
\lambda
\equiv
\lambda_{h_0^i}(t),
\qquad
\pi
\equiv
\pi_i(t)
=
\sum_{k=1}^{K}
g_k^i(t)\lambda_k(t),
\label{eq:rates}
\end{equation}
where $\lambda_{h_0^i}(t)$ is the gold reveal probability associated with the true
hierarchy class $h_0^i$, and $\lambda_k(t)$ is the (schedule-determined)
reveal probability of class $k$. The quantity $\pi_i(t)$ is the model's total
predicted reveal probability, obtained by averaging the class-dependent reveal
probability $\lambda_k(t)$ under the Pass-1 distribution $g^i(t)$.

\paragraph{Forward posterior at masked positions.}
At an unmasked position ($s_t^i \neq s_m$) we have $s_{t-1}^i = s_0^i$
deterministically, contributing zero KL. At a masked position the
absorbing-state posterior is
\begin{equation}
  q(s_{t-1}^i \mid s_t^i, s_0^i) =
  \begin{cases}
    \mathrm{Cat}\bigl(s_{t-1}^i;\, s_t^i\bigr), & s_t^i \neq \mathbf{s_m}, \\[8pt]
    \mathrm{Cat}\!\left(
      s_{t-1}^i;\,
      \dfrac{(1-\alpha_{h_0^i}(t-1))\,\mathbf{s_m}
             + (\alpha_{h_0^i}(t-1) - \alpha_{h_0^i}(t))\,s_0^i}
            {1 - \alpha_{h_0^i}(t)}
    \right), & s_t^i = \mathbf{s_m}.
  \end{cases}
  \label{eq:reverse_joint}
\end{equation}

\paragraph{Reverse hierarchy model (Pass 1).}

The head first assigns position $i$ to
class $h$ with probability $g_h^i(t)$, and each class then reveals at its own
\emph{known} schedule probability $\lambda_h(t)$. The reverse model is thus
\begin{equation}
p_\theta(h_{t-1}^i = h
\mid
\mathbf{x}_t,\mathbf{h}_t)
=
\lambda_h(t)\,g_h^i(t),
\qquad
h\in\{1,\ldots,K\},
\label{eq:h_reverse}
\end{equation}
\begin{equation}
p_\theta(h_{t-1}^i = m
\mid
\mathbf{x}_t,\mathbf{h}_t)
=
1-\pi,
\qquad
\pi_i(t)
=
\sum_{k=1}^{K}
g_k^i(t)\lambda_k(t).
\label{eq:h_reverse_mask}
\end{equation}
Each
class here reveals at its own probability $\lambda_h(t)$, and on the gold class this
gives the \emph{gold} probability directly,
\begin{equation}
p_\theta(h_{t-1}^i = h_0^i\mid \mathbf{x}_t,\mathbf{h}_t)
=
\lambda_{h_0^i}(t)\,g_{h_0^i}^i(t)
=
\lambda\,g_{h_0^i}^i(t).
\label{eq:h_reverse_gold}
\end{equation}

\paragraph{Per-position KL.}

\[
\renewcommand{\arraystretch}{1.4}
\begin{array}{lcc}
\toprule
h_{t-1}^i
&
q
&
p_\theta
\\
\midrule
h_0^i
&
\lambda
&
\lambda\,g_{h_0^i}^i(t)
\\
h\neq h_0^i
&
0
&
\lambda_h(t)\,g_h^i(t)
\\
h_{\mathrm{mask}}
&
1-\lambda
&
1-\pi
\\
\bottomrule
\end{array}
\]

The KL divergence is
\begin{equation}
D_{\mathrm{KL}}(q\|p_\theta)
=
\sum_v
q(v)
\log
\frac{q(v)}{p_\theta(v)},
\qquad
v\in\{1,\ldots,K,h_{\mathrm{mask}}\}.
\label{eq:kl_def}
\end{equation}
Since $q(v)=0$ for every incorrect hierarchy class ($h\neq h_0^i$), all such terms
vanish --- the off-gold entries $\lambda_h(t)\,g_h^i(t)$ never enter the KL and
serve only to normalize $p_\theta$. Only the gold-class and mask rows remain,
\begin{equation}
D_{\mathrm{KL}}
=
\lambda
\log
\frac{\lambda}
{\lambda\,g_{h_0^i}^i(t)}
+
(1-\lambda)
\log
\frac{1-\lambda}
{1-\pi}.
\label{eq:kl_two_rows}
\end{equation}
In the gold-class row the reveal rate $\lambda$ cancels between numerator and
denominator, leaving a pure cross-entropy on $g^i$:
\begin{align}
D_{\mathrm{KL}}
&=
\lambda
\log
\frac{1}{g_{h_0^i}^i(t)}
+
(1-\lambda)
\log
\frac{1-\lambda}{1-\pi}
\nonumber\\[4pt]
&=
-\lambda
\log g_{h_0^i}^i(t)
+
(1-\lambda)
\log
\frac{1-\lambda}{1-\pi}.
\label{eq:kl_final_intermediate}
\end{align}
Hence
\begin{equation}
{\;
D_{\mathrm{KL}}^{\mathrm{hier,i}}
=
\underbrace{-\lambda\,\log g_{h_0^i}^i(t)}_{\text{reveal-weighted cross-entropy}}
\;+\;
\underbrace{(1-\lambda)\log\dfrac{1-\lambda}{1-\pi}}_{\text{extra-term}}
\;}
\label{eq:kl_final}
\end{equation}
Summing over all the positions,
\begin{equation}
{\;
L_t^{\mathrm{hier}}
=
\sum_{i=1}^{L}
\mathbb{E}_q\!\left[
-\lambda_{h_0^i}(t)\,\log g_{h_0^i}^i(t)
\;+\;
\big(1-\lambda_{h_0^i}(t)\big)\log\frac{1-\lambda_{h_0^i}(t)}{1-\pi_i(t)}
\right].
\;}
\label{eq:Lt_hier_final}
\end{equation}


\subsection{Step 9 — Expand the Token KL (Pass 2)}
\label{sec:token-kl-expansion}

We expand the per-position token contribution
\begin{equation}
L^{\mathrm{tok},i}_t
=\mathbb{E}_{q(h^i_{t-1}\mid \mathbf{x}_t,\mathbf{h}_t,\mathbf{x}_0,\mathbf{h}_0)}
\Big[\,
D_{\mathrm{KL}}\!\big(
q(x^i_{t-1}\mid h^i_{t-1},\cdots,\mathbf{x}_0,\mathbf{h}_0)
\,\big\|\,
p_\theta(x^i_{t-1}\mid h^i_{t-1},\mathbf{x}_t)
\big)
\Big],
\label{eq:tok-i-start}
\end{equation}

By the shared reveal variable (~\eqref{eq:forward_bernoulli}), position $i$
is governed by a single Bernoulli $b^i_t\sim\mathrm{Bernoulli}\!\big(1-\alpha_{h^i_0}(t)\big)$,
and the joint state $s^i_t=(x^i_t,h^i_t)$ is a deterministic push-forward
of $b^i_t$: both streams reveal to $(x^i_0,h^i_0)$ when the position is
kept and mask to $(m,h_{\mathrm{mask}})$ otherwise. Consequently the joint
reverse posterior over $(x^i_{t-1},h^i_{t-1})$ is~\eqref{eq:reverse_joint}.

\paragraph{Branch decomposition.}
The outer expectation in \eqref{eq:tok-i-start} runs over the two values
$h^i_{t-1}\in\{h^i_0,\,h_{\mathrm{mask}}\}$, with marginal weights
$\lambda$ and $1-\lambda$ from \eqref{eq:reverse_joint}. We treat each branch
in turn.

\textbf{\emph{Revealed branch}} ($h^i_{t-1}=h^i_0$, weight $\lambda$).

$p_\theta(x^i_{t-1}=v\mid h^i_0,x_t,h_{t-1})=f^i_v(t)$ with
$f^i_v(t):=[x_\theta(x_t,h_{t-1},t)]_v$ and $\sum_{v=1}^V f^i_v(t)=1$:
\begin{equation}
\renewcommand{\arraystretch}{1.25}
\begin{array}{c|c|c}
x^i_{t-1} & q & p_\theta \\ \hline
x^i_0        & 1 & f^i_{x^i_0}(t) \\
v\neq x^i_0  & 0 & f^i_v(t) \\
m            & 0 & 0
\end{array}
\qquad\Longrightarrow\qquad
D_{\mathrm{KL}}\big(q\,\|\,p_\theta\big)\big|_{\text{rev}}
=\sum_{v} q_v\log\frac{q_v}{p_{\theta,v}}
=-\log f^i_{x^i_0}(t).
\label{eq:branch-A}
\end{equation}

\textbf{\emph{Masked branch}} ($h^i_{t-1}=h_{\mathrm{mask}}$, weight $1-\lambda$).

\begin{equation}
\renewcommand{\arraystretch}{1.25}
\begin{array}{c|c|c}
x^i_{t-1} & q & p_\theta \\ \hline
x^i_0        & 0 & 0 \\
v\neq x^i_0  & 0 & 0 \\
m            & 1 & 1
\end{array}
\qquad\Longrightarrow\qquad
D_{\mathrm{KL}}\big(q\,\|\,p_\theta\big)\big|_{\text{mask}}=0.
\label{eq:branch-B}
\end{equation}

{Taking the outer expectation.}

\begin{align}
L^{\mathrm{tok},i}_t
&=\mathbb{E}_{q}\!\Big[
\underbrace{\lambda}_{P(h^i_{t-1}=h^i_0)}
\cdot\big(-\log f^i_{x^i_0}(t)\big)
\;+\;
\underbrace{(1-\lambda)}_{P(h^i_{t-1}=h_{\mathrm{mask}})}
\cdot 0
\Big]
\nonumber\\[2pt]
&=-\,\mathbb{E}_{q}\!\big[\,\lambda_{h^i_0}(t)\,\log f^i_{x^i_0}(t)\big].
\label{eq:tok-i-final}
\end{align}

Summing over position, 
\begin{equation}
{\;
L^{\mathrm{tok}}_t
=\sum_{i=1}^{L}
L^{\mathrm{tok},i}_t
=-\sum_{i=1}^{L}
\mathbb{E}_{q}\!\big[\,\lambda_{h^i_0}(t)\,\log f^i_{x^i_0}(t)\big].
\;}
\label{eq:tok-final}
\end{equation}
The token loss is thus a \emph{reveal-weighted cross-entropy} on the
Pass-2 softmax: each position contributes the negative log-likelihood of
its clean token $x^i_0$, weighted by the class-dependent probability
$\lambda_{h^i_0}(t)$ that the position is revealed at step $t-1$.

\section*{Discrete $\to$ Continuous}

\textbf{Setup.} Let $s=t-\tfrac{1}{T}$, and let $\alpha_k(t)$ be the class-$k$ survival
(unmasked) probability, decreasing with $\alpha_k(0)=1$, $\alpha_k(1)=0$.

\textbf{Step A --- per-step masked-position KL}
\begin{equation}
L^{\mathrm{tok},i}_{t}
=-\,\frac{\alpha_{h^i_0}(s)-\alpha_{h^i_0}(t)}{1-\alpha_{h^i_0}(t)}\,
\log f^i_{x^i_0}(t).
\end{equation}

\textbf{Step B --- sum over timesteps.}
\begin{equation}
\mathcal{L}^{\mathrm{tok}}
=\sum_{t} L^{\mathrm{tok}}_t
=-\sum_{t}\sum_i \mathbb{E}_q\!\left[
\frac{\alpha_{h^i_0}(s)-\alpha_{h^i_0}(t)}{1-\alpha_{h^i_0}(t)}\,
\log f^i_{x^i_0}(t)\right].
\end{equation}

\textbf{Step C --- expose the mask indicator.}
\begin{equation}
\mathcal{L}^{\mathrm{tok}}
=-\sum_{t}\mathbb{E}_q\!\left[
\sum_i {1}[x^i_t=M]\,
\frac{\alpha_{h^i_0}(s)-\alpha_{h^i_0}(t)}{1-\alpha_{h^i_0}(t)}\,
\log f^i_{x^i_0}(t)\right].
\end{equation}

\textbf{Step D --- Riemann rescaling ($\sum_t \to \mathbb{E}_t$).}
\begin{equation}
\sum_{t}[\,\cdot\,]
=\underbrace{\tfrac1T\sum_{t}}_{\mathbb{E}_{t\in\{1/T,\dots,1\}}}
\underbrace{T[\,\cdot\,]}_{O(1)},
\end{equation}

\begin{equation}
\mathcal{L}^{\mathrm{tok}}
=-\,\mathbb{E}_{t}\,\mathbb{E}_q\!\left[
\sum_i {1}[x^i_t=M]\,
\Big(T\cdot\tfrac{\alpha_{h^i_0}(s)-\alpha_{h^i_0}(t)}{1-\alpha_{h^i_0}(t)}\Big)
\log f^i_{x^i_0}(t)\right].
\end{equation}

\textbf{Step E --- limit $T\to\infty$: weight $\to$ reveal hazard.}
\begin{equation}
T\big(\alpha_{h^i_0}(s)-\alpha_{h^i_0}(t)\big)
=\frac{\alpha_{h^i_0}(t-\tfrac1T)-\alpha_{h^i_0}(t)}{1/T}
\;\xrightarrow[T\to\infty]{}\; -\,\alpha'_{h^i_0}(t),
\end{equation}

\begin{equation}
T\cdot\frac{\alpha_{h^i_0}(s)-\alpha_{h^i_0}(t)}{1-\alpha_{h^i_0}(t)}
\;\longrightarrow\;
\frac{-\alpha'_{h^i_0}(t)}{1-\alpha_{h^i_0}(t)}
=:\lambda'_{h^i_0}(t).
\end{equation}

\textbf{Step F --- continuous-time loss.}
\begin{equation}
{\;
\mathcal{L}_{\mathrm{LM}}
=\mathbb{E}_{t,\,q}\!\left[
\sum_i {1}[x^i_t=M]\,\lambda'_{h^i_0}(t)\,
\big(-\log f^i_{x^i_0}(t)\big)\right].}
\tag{61}
\end{equation}

\textbf{Step G --- Hierarchy head (same derivation, $f\to g$).}
\begin{equation}
\mathcal{L}_{\mathrm{hier}}
=\mathbb{E}_{t,\,q}\!\left[
\sum_i {1}[x^i_t=M]\,\lambda'_{h^i_0}(t)\,
\big(-\log g^i_{h^i_0}(t)\big)\right].
\tag{62}
\end{equation}

Let $s = t - \tfrac{1}{T}$,  and 
\begin{equation}
  \mathcal{E}^{i}_t
  = \bigl(1-\lambda_{h^i_0}(t)\bigr)\,
    \log\frac{1-\lambda_{h^i_0}(t)}{1-\pi_i(t)}, 
\end{equation}
Both $\lambda_{h^i_0}(t)$ and every $\lambda_k(t)$ are $O(1/T)$, hence so is
$\pi_i(t)$.

\paragraph{Step H --- expose the mask indicator.}
Only masked positions contribute (unmasked positions give zero KL, cf.\ ~\eqref{eq:tok-i-start}):
\begin{equation}
  \mathcal{E}^{\mathrm{hier}}_t
  = \mathbb{E}_q\!\left[\sum_i {1}[x^i_t = M]\,
    \bigl(1-\lambda_{h^i_0}(t)\bigr)
    \log\frac{1-\lambda_{h^i_0}(t)}{1-\pi_i(t)}\right].
\end{equation}

\paragraph{Step I --- small-step expansion.}
Write $\lambda \equiv \lambda_{h^i_0}(t)$ and $\pi \equiv \pi_i(t)$, both $O(1/T)$.
Using $\log(1-z) = -z - \tfrac{z^2}{2} - \cdots$,
\begin{equation}
  \log\frac{1-\lambda}{1-\pi}
  = (\pi-\lambda) - \tfrac12\bigl(\lambda^2-\pi^2\bigr) + O(T^{-3}).
\end{equation}
Multiplying by $(1-\lambda)$ and collecting terms by order,
\begin{equation}
  \mathcal{E}^i_t
  = \underbrace{(\pi-\lambda)}_{O(1/T)}
  \;+\; \underbrace{\tfrac12(\lambda-\pi)^2}_{O(1/T^2)}
  \;+\; O(T^{-3}).
\end{equation}

\textbf{{Step J --- Riemann rescaling ($\sum_t \to \mathbb{E}_t$).}}
\begin{equation}
  \sum_t [\,\cdot\,]
  = \underbrace{\frac{1}{T}\sum_t}_{\mathbb{E}_{t\in\{1/T,\dots,1\}}}
    \underbrace{T\,[\,\cdot\,]}_{O(1)},
  \qquad
  T\,\mathcal{E}^i_t
  = \underbrace{T(\pi-\lambda)}_{O(1)}
  + \underbrace{\tfrac{T}{2}(\lambda-\pi)^2}_{O(1/T)\;\to\;0}.
\end{equation}

\textbf{{Step K --- limit $T\to\infty$: rate $\to$ reveal hazard.}}
For each class $k$, as in Step~E of the token derivation,
\begin{equation}
  T\,\lambda_k(t)
  = \frac{\alpha_k\!\left(t-\tfrac1T\right)-\alpha_k(t)}{1/T}\cdot
    \frac{1}{1-\alpha_k(t)}
  \;\xrightarrow[T\to\infty]{}\;
  \frac{-\alpha_k'(t)}{1-\alpha_k(t)} =: \lambda_k'(t).
\end{equation}
Applying this to $\lambda$ and, term by term, to $\pi$,
\begin{equation}
  T\lambda = T\lambda_{h^i_0}(t) \to \lambda_{h^i_0}'(t),
  \qquad
  T\pi = \sum_k g^i_k(t)\,T\lambda_k(t) \to \sum_k g^i_k(t)\,\lambda_k'(t),
\end{equation}
so that
\begin{equation}
  T(\pi-\lambda) \;\xrightarrow[T\to\infty]{}\;
  \sum_k g^i_k(t)\,\lambda_k'(t) - \lambda_{h^i_0}'(t),
\end{equation}
while the quadratic remainder vanishes.

Now, the $1-\lambda_k(t)$ term, 
\begin{equation}
1-\lambda_k(t) = \frac{1-\alpha_k\!\left(t-\tfrac{1}{T}\right)}{1-\alpha_k(t)} .
\end{equation}
Since $\alpha_k$ is continuous, $\alpha_k\!\left(t-\tfrac{1}{T}\right)\to\alpha_k(t)$ as $T\to\infty$.
Moreover, $t>0$ implies $1-\alpha_k(t)>0$, so the denominator is nonzero. Therefore
\begin{equation}
1-\lambda_k(t) \;\xrightarrow[T\to\infty]{}\; \frac{1-\alpha_k(t)}{1-\alpha_k(t)} = 1 .
\end{equation}

The only singular point is $t=0$: since $1-\alpha_k(0)=0$,
\begin{equation}
\lambda'_k(t)\xrightarrow[t\to 0^+]{}\infty ,
\end{equation}
while $1-\lambda_k(t)\in(0,1]$ remains bounded and is undefined only at $t=0$ itself.

In pratice, we sample $t\sim\mathcal{U}[\varepsilon,1]$, and therefore never sample $t=0$ and hence the expression always stays bounded, where $\varepsilon = 10^{-3}$.

\textbf{{Step L --- continuous-time extra term.}}
\begin{equation}
  {\;
  \mathcal{L}_{\mathrm{cons}}
  = \mathbb{E}_{t,\,q}\!\left[\sum_i {1}[x^i_t = M]
    \Bigl(
      \underbrace{\textstyle\sum_{k} g^i_k(t)\,\lambda_k'(t)}
      _{\text{predicted reveal hazard}}
      \;-\; \lambda_{h^i_0}'(t)
    \Bigr)\right],
  \qquad
  \lambda_k'(t) = \frac{-\alpha_k'(t)}{1-\alpha_k(t)}.
  \;}
\end{equation}

\subsection{Step 10 --- Final Training Loss}

\paragraph{Training losses.}
The final training loss consists of the following three components:
\begin{align}
  \mathcal{L}_{\mathrm{LM}}
  &= \mathbb{E}_{t,q}\!\left[
       \sum_{i=1}^{L}
       \mathbf{1}[x_t^i = M]\, \lambda'_{h_0^i}(t)\,
       \bigl(-\log f_{x_0^i}^i(t)\bigr)
     \right],
  \label{eq:loss_lm}\\[4pt]
  \mathcal{L}_{\mathrm{hier}}
  &= \mathbb{E}_{t,q}\!\left[
       \sum_{i=1}^{L}
       \mathbf{1}[x_t^i = M]\, \lambda'_{h_0^i}(t)\,
       \bigl(-\log g_{h_0^i}^i(t)\bigr)
     \right],
  \label{eq:loss_hier}\\[4pt]
  \mathcal{L}_{\mathrm{cons}}
  &= \mathbb{E}_{t,\,q}\!\left[
       \sum_{i=1}^{L}
       \mathbf{1}[x_t^i = M]\,
       \Bigl(
         {\textstyle\sum_{k} g^i_k(t)\,\lambda_k'(t)}
         \;-\; \lambda_{h_0^i}'(t)
       \Bigr)
     \right],
  \label{eq:loss_cons}
\end{align}

In practice we optimise a reweighted sum of the same
three terms, replacing the tied weights above by tunable
coefficients:
\begin{equation}
  {\;
  \mathcal{L}_{\mathrm{HMDLM}}
  = \mathcal{L}_{\mathrm{LM}}
    \;+\; \beta_{\mathrm{hier}}\,\mathcal{L}_{\mathrm{hier}}
    \;+\; \beta_{\mathrm{cons}}\,\mathcal{L}_{\mathrm{cons}}.
  \;}
  \label{eq:total_loss}
\end{equation}

\begin{algorithm}[t]
\caption{HMDLM training\protect\footnotemark}
\label{alg:hmdlm_train}
\begin{algorithmic}[1]
\Require data $\mathcal{D}$ of clean pairs $(\mathbf{x}_0,\mathbf{h}_0)$, with
$x_{0,i}\!\in\!\{1,\dots,V\}$, $h^i_{0}\!\in\!\{1,\dots,K\}$;
per-class keep schedules $\{\alpha_k(\cdot)\}$ with reveal probabilities $\lambda_k(\cdot)$;
hierarchy head $h_\theta$ (Pass 1), token head $x_\theta$ (Pass 2);
number of inference time steps $T$, weights $\beta_{\mathrm{hier}},\beta_{\mathrm{cons}}$; learning rate $\eta$.
\Statex
\While{not converged}
  \State sample $(\mathbf{x}_0,\mathbf{h}_0)\sim\mathcal{D}$,\quad $t\sim\mathcal{U}(\epsilon,1)$
  \Statex \hspace{\algorithmicindent}\textit{// Forward corruption to level $t$ }
  \For{$i=1,\dots,L$}
    \State $m_i \sim \mathrm{Bernoulli}\!\bigl(1-\alpha_{h^i_{0}}(t)\bigr)$
      \Comment{$m_i\!=\!1$: masked at $t$}
    \State $(x^i_{t},h^i_{t}) \gets (M,\,h_{\mathrm{mask}})$ if $m_i\!=\!1$ else $(x^i_{0},\,h^i_{0})$
  \EndFor
  \Statex \hspace{\algorithmicindent}\textbf{Pass 1 --- hierarchy.} 
  \State $g_i(t) \gets \mathrm{softmax}\bigl(h^i_{\theta}(\mathbf{x}_t,\mathbf{h}_t,t)\bigr)$,\quad
         $\lambda'_i \gets \lambda'_{h_{0}^i}(t)$, \quad for all $i$
  \State $\mathcal{L}_{\mathrm{hier}} \gets
      \sum_{i} m_i\,\lambda'_i\,\bigl(-\log g^i_{h^i_{0}}(t)\bigr)$
  \State $\mathcal{L}_{\mathrm{cons}} \gets
      \sum_{i} m_i\,({\textstyle\sum_{k} g^i_k(t)\,\lambda_k'(t)}
         \;-\; \lambda_{h_0^i}'(t))$
  \Statex \hspace{\algorithmicindent}
  \State $s \gets \max(t - 1/T,\, 0)$
\Comment{one reverse step on the inference grid}
\For{$i$ with $m_i = 1$}
    \State $\lambda_i \gets
      \begin{cases}
        \dfrac{\alpha_{h_0^i}(s) - \alpha_{h_0^i}(t)}{1 - \alpha_{h_0^i}(t)}, & \alpha_{h_0^i}(t) < 1,\\[6pt]
        0, & \text{otherwise}
      \end{cases}$
    \Comment{per-step reveal probability}
    \State $b_i \sim \mathrm{Bernoulli}(\lambda_i)$
    \Comment{$b_i = 1$: hierarchy revealed at $s$}
\EndFor
\State $\mathcal{R} \gets \{\, i : m_i = 1,\ b_i = 1 \,\}$
\State $h_s^i \gets h_0^i$ for $i \in \mathcal{R}$;\ \ $h_s^i \gets h_t^i$ otherwise
\Comment{Pass-2 context only: gold hierarchy at $\mathcal{R}$, $h_{\text{mask}}$ elsewhere}
  \Statex \hspace{\algorithmicindent}\textbf{Pass 2 --- token.} 
  \State $f_i(t) \gets \mathrm{softmax}\bigl(x^i_{\theta}(\mathbf{x}_t,\,\mathbf{h}_{s},\,t)\bigr)$\quad for all $i$
  \State $\mathcal{L}_{\mathrm{LM}} \gets \sum_{i} m_i\,\lambda'_i\,\bigl(-\log f^i_{x^i_{0}}(t)\bigr)$
    \Comment{closed-form token CE over \emph{all} masked positions }
  \Statex \hspace{\algorithmicindent}\textit{// Combine and update}
  \State $\mathcal{L} \gets
      (\mathcal{L}_{\mathrm{LM}}
      + \beta_{\mathrm{hier}}\,\mathcal{L}_{\mathrm{hier}}
      + \beta_{\mathrm{cons}}\,\mathcal{L}_{\mathrm{cons}})/L$
  \State $\theta \gets \theta - \eta\,\nabla_\theta \mathcal{L}$
\EndWhile
\end{algorithmic}
\end{algorithm}
\footnotetext{We tried training the second stage with all gold hierarchy. However, the performance during inference was not as expected since the model never learned to predict the correct token distribution without gold hierarchy, which is the case during inference (inference sees masked hierarchy during second stage). Therefore, we added steps 10 - 16 where in we sample token positions and pass gold hierarchy during training only for this subset, and rest of the positions mask their hierarchy.}
\begin{algorithm}[t]
\caption{HMDLM ancestral sampling (reverse process)}
\label{alg:hmdlm_sample}
\begin{algorithmic}[1]
\Require hierarchy head $h_\theta$ (Pass 1), token head $x_\theta$ (Pass 2);
per-class keep schedules $\{\alpha_k(\cdot)\}_{k=1}^{K}$; number of steps $T$;
optional token temperature $\tau$.
\Statex
\State $x^i \gets M,\quad h^i \gets h_{\mathrm{mask}}$ \quad for all $i=1,\dots,L$
  \Comment{$t=1$: fully masked}
\For{$n = T,\,T-1,\,\dots,\,1$}
  \State $t \gets n/T$,\qquad $s \gets (n-1)/T$
    \Comment{reverse step $t\!\to\!s$; $\mathbf{x},\mathbf{h}$ persist across iterations}
  \Statex \hspace{\algorithmicindent}\textbf{Pass 1 --- hierarchy.} 
  \State $g^i \gets \mathrm{softmax}\bigl(h_\theta^i(\mathbf{x},\mathbf{h},t)\bigr)$ \quad for all $i$ with $h^i=h_{\mathrm{mask}}$
  \State $\mathcal{R} \gets \varnothing$
  \For{$i$ with $h^i = h_{\mathrm{mask}}$}
    \State $\lambda_k \gets \dfrac{\alpha_k(s)-\alpha_k(t)}{1-\alpha_k(t)}$ \quad for $k=1,\dots,K$
      \Comment{per-step reveal probability of class $k$}
    \State sample $\hat h^i \sim \mathrm{Cat}\!\bigl(\lambda_1 g_1^i,\ \dots,\ \lambda_K g_K^i,\ \underbrace{1-\textstyle\sum_k \lambda_k g_k^i}_{\text{stay masked}}\bigr)$;\quad $h^i \gets \hat h^i$
      \Comment{commit the sampled class (or $h_{\mathrm{mask}}$) into $\mathbf{h}$}
    \If{$h^i \neq h_{\mathrm{mask}}$}
      \State $\mathcal{R} \gets \mathcal{R}\cup\{i\}$
        \Comment{position revealed this step (shared across both streams)}
    \EndIf
  \EndFor
  \Statex \hspace{\algorithmicindent}\textbf{Pass 2 --- token.} \textit{Fill tokens at $\mathcal{R}$, conditioned on the \textbf{just-sampled} hierarchy $\hat h^i$ .}
  \State $f^i \gets \mathrm{softmax}\bigl(x_\theta^i(\mathbf{x},\mathbf{h},t))$ \quad for all $i\in\mathcal{R}$
    \Comment{$\mathbf{h}$ carries $\hat h^i$ at $i\in\mathcal{R}$ from line~9}
  \For{$i \in \mathcal{R}$}
    \State sample $x^i \sim \mathrm{Cat}(f^i)$
      \Comment{reveal token; masked positions keep $x^i=M$}
  \EndFor
  \State \textbf{freeze} $(x^i,h^i)$ for all $i\in\mathcal{R}$; carry all of $(\mathbf{x},\mathbf{h})$ into step $n-1$
    \Comment{committed positions are never revisited}
\EndFor
\State \Return $(\mathbf{x},\mathbf{h})$
  \Comment{clean sample $(\mathbf{x}_0,\mathbf{h}_0)$}
\end{algorithmic}
\end{algorithm}

\section{Expected Mask Fraction under Hierarchical Log-Linear Schedule}
\label{sec:expected_mask_fraction}

We derive the expected fraction of masked tokens per training step under our
hierarchical masking schedule, which is needed to compute the effective
training-token budget and compare fairly against baseline diffusion language
models~\citep{sahoo2024simple}.

\paragraph{Setup.}
Let $K$ denote the number of hierarchy levels and let $\text{offset}_k \in [0, 1)$
denote the per-level time offset. The effective time at level $k$ is
\begin{equation}
    t_{\text{eff},k}
    \;=\; \mathrm{clip}\!\left(
        \frac{t - \text{offset}_k}{1 - \text{offset}_k},\; 0,\; 1
    \right),
    \label{eq:teff}
\end{equation}
and the log-linear schedule for level $k$ is
\begin{equation}
    \alpha_k(t) \;=\; 1 - (1 - \varepsilon)\, t_{\text{eff},k},
    \qquad \varepsilon = 10^{-3}.
    \label{eq:alpha_k}
\end{equation}
The probability that a token at level $k$ is masked at time $t$ is therefore
\begin{equation}
    p_k(t) \;=\; 1 - \alpha_k(t) \;=\; (1 - \varepsilon)\, t_{\text{eff},k}.
    \label{eq:pk}
\end{equation}

\paragraph{Per-level expected mask probability.}
We compute $\mathbb{E}_{t \sim \mathcal{U}[0,1]}[p_k(t)]
= (1-\varepsilon)\,\mathbb{E}[t_{\text{eff},k}]$.
Splitting the integral at the clamp boundary $t = \text{offset}_k$:
\begin{align}
    \mathbb{E}[t_{\text{eff},k}]
    &= \int_0^{\text{offset}_k} 0\, dt
       + \int_{\text{offset}_k}^{1}
            \frac{t - \text{offset}_k}{1 - \text{offset}_k}\, dt.
\end{align}
With the substitution $u = t - \text{offset}_k$, the second integral becomes
\begin{equation}
    \int_0^{\,1 - \text{offset}_k}
        \frac{u}{1 - \text{offset}_k}\, du
    \;=\; \frac{(1 - \text{offset}_k)^2}{2(1 - \text{offset}_k)}
    \;=\; \frac{1 - \text{offset}_k}{2}.
\end{equation}
Hence,
\begin{equation}
    {\;
    \mathbb{E}_{t \sim \mathcal{U}[0,1]}[\, p_k(t)\,]
    \;=\; (1 - \varepsilon) \cdot \frac{1 - \text{offset}_k}{2}.
    \;}
    \label{eq:per_level_expectation}
\end{equation}

\paragraph{Plugging in our configuration.}
With $K = 2$, level offsets $\text{offset}_1 = 0.25$ (rare tokens) and
$\text{offset}_0 = 0.0$ (frequent tokens), and $\varepsilon = 10^{-3}$:
\begin{align}
    \mathbb{E}[p_1(t)] &= 0.999 \cdot \tfrac{1 - 0.25}{2}
                       = 0.999 \cdot 0.375
                       = 0.374625, \\
    \mathbb{E}[p_0(t)] &= 0.999 \cdot \tfrac{1 - 0.0}{2}
                       = 0.999 \cdot 0.500
                       = 0.499500.
\end{align}

\paragraph{Overall expected mask fraction.}
Let $f_k$ denote the fraction of \emph{real} token positions assigned to level
$k$, with $f_0 + f_1 = 1$. The overall expected mask fraction is
\begin{equation}
    \mathbb{E}[\text{mask fraction}]
    \;=\; f_0 \cdot \mathbb{E}[p_0(t)]
        + f_1 \cdot \mathbb{E}[p_1(t)].
    \label{eq:overall_mask_fraction}
\end{equation}
For LM1B with the \texttt{bert-base-uncased} tokenizer and context length
$L = 128$, we assign a token to level $1$ if it occurs at most $\tau$ times
within its own sequence (at $\tau = 1$, exactly once). Measured over the full
training split ($30{,}301{,}028$ sequences, $922{,}347{,}589$ real tokens),
level~$1$ holds $737{,}634{,}266$ real tokens at $\tau = 1$ and
$914{,}638{,}961$ at $\tau = 5$, so
\begin{align}
    \tau = 1:&\quad f_1 = 0.799736, \quad f_0 = 0.200264, \\
    \tau = 5:&\quad f_1 = 0.991642, \quad f_0 = 0.008358.
\end{align}
Substituting into \eqref{eq:overall_mask_fraction}:
\begin{align}
    \tau=1:\quad
    \mathbb{E}[\text{mask fraction}]
    &= 0.200264(0.499500)
      + 0.799736(0.374625) \nonumber\\
    &= 0.100032 + 0.299601
     \approx 0.399633, \\
    \tau=5:\quad
    \mathbb{E}[\text{mask fraction}]
    &= 0.008358(0.499500)
      + 0.991642(0.374625) \nonumber\\
    &= 0.004175 + 0.371494
     \approx 0.375669.
\end{align}

\paragraph{Effective training-token budget.}
The budget is denominated in \emph{supervised} positions, i.e.\ those that are
both masked and real. Padding is excluded from the loss.\footnote{LM1B sequences are not packed, so a $128$-position row holds
$\bar{R} = 922{,}347{,}589 / 30{,}301{,}028 = 30.44$ real tokens on average and
the remainder is padding.} Given training steps $S$, batch size $B$, and mean real length
$\bar{R}$, the expected number of supervised tokens is
\begin{equation}
    \mathbb{E}[\text{supervised tokens}]
    \;=\; S \cdot B \cdot \bar{R} \cdot \mathbb{E}[\text{mask fraction}].
    \label{eq:effective_tokens}
\end{equation}
With $B = 512$ and $\bar{R} = 30.44$, each step presents
$B \cdot \bar{R} = 15{,}585$ real positions and therefore supervises
$15{,}585 \cdot 0.399633 \approx 6{,}228.3$ tokens at $\tau = 1$ and
$15{,}585 \cdot 0.375669 \approx 5{,}854.8$ at $\tau = 5$, against MDLM's
$15{,}585 \cdot 0.499500 \approx 7{,}784.7$. To reach a target budget $N$, the
required number of steps is
\begin{equation}
    S^{\star}
    \;=\; \frac{N}{B \cdot \bar{R} \cdot \mathbb{E}[\text{mask fraction}]}.
    \label{eq:required_steps}
\end{equation}
We instantiate this at two budgets. MDLM's $33$B tokens on LM1B counts
masked positions over the padded context ($10^{6} \cdot 65{,}536 \cdot 0.4995 =
32.74$B); of these, a fraction $\bar{R}/L = 0.2378$ are real, so the same run
supervises $N_{33} = 7.7847$B tokens. Matching it requires
\begin{align}
    S^{\star}_{33}
    &= \frac{7.7847 \times 10^{9}}{15{,}585 \cdot 0.399633}
     \approx 1.2499 \times 10^{6} \text{ steps} && (\tau = 1), \\
    S^{\star}_{33}
    &= \frac{7.7847 \times 10^{9}}{15{,}585 \cdot 0.375669}
     \approx 1.3296 \times 10^{6} \text{ steps} && (\tau = 5).
\end{align}
For the extended budget, twice as large at $N_{65} = 15.5694$B supervised
tokens ($65.47$B masked positions),
\begin{align}
    S^{\star}_{65}
    &= \frac{15.5694 \times 10^{9}}{15{,}585 \cdot 0.399633}
     \approx 2.4998 \times 10^{6} \text{ steps} && (\tau = 1), \\
    S^{\star}_{65}
    &= \frac{15.5694 \times 10^{9}}{15{,}585 \cdot 0.375669}
     \approx 2.6593 \times 10^{6} \text{ steps} && (\tau = 5).
\end{align}

\section{Expected MAsking Fraction for WikiHow}
\label{app:exp_masking_frac_wikihow}
Under the hierarchical power schedule, a response token at level $\ell$ with
exponent $w_\ell$ is masked at diffusion time $t \sim \mathcal{U}(0,1)$ with
probability $(1-\epsilon)\,t^{w_\ell}$. Its expected masking rate is therefore
\begin{equation}
  \bar{m}_\ell \;=\; (1-\epsilon)\int_0^1 t^{w_\ell}\,dt \;=\; \frac{1-\epsilon}{w_\ell+1}.
\end{equation}
Let $g_\ell$ be the fraction of tokens assigned to level $\ell$, pooled over the
training set and counted over diffused positions only (the prompt and padding
are excluded). The expected masking rate of HMDLM is then
\begin{equation}
  \bar{m}_{\text{HMDLM}} \;=\; (1-\epsilon)\sum_{\ell} \frac{g_\ell}{w_\ell+1}.
\end{equation}
MDLM corresponds to a single level with $w=1$, giving
$\bar{m}_{\text{MDLM}} = (1-\epsilon)/2$. With two levels and exponents
$w_0 = 1$, $w_1 = 2$,
\begin{equation}
  \bar{m}_{\text{HMDLM}} \;=\; (1-\epsilon)\left[\frac{1-g_1}{2} + \frac{g_1}{3}\right]
  \;=\; (1-\epsilon)\left(\frac{1}{2} - \frac{g_1}{6}\right),
\end{equation}
so the ratio to MDLM, in which $\epsilon$ cancels, is
\begin{equation}
  \rho \;=\; \frac{\bar{m}_{\text{HMDLM}}}{\bar{m}_{\text{MDLM}}} \;=\; 1 - \frac{g_1}{3}.
\end{equation}
To match the expected number of masked (loss-bearing) tokens seen by MDLM in
$S_{\text{MDLM}}$ steps at the same batch size, HMDLM is trained for
\begin{equation}
  S_{\text{HMDLM}} \;=\; \frac{S_{\text{MDLM}}}{\rho} \;=\; \frac{3\,S_{\text{MDLM}}}{3 - g_1},
  \qquad
  \Delta S \;=\; S_{\text{HMDLM}} - S_{\text{MDLM}} \;=\; S_{\text{MDLM}}\,\frac{g_1}{3 - g_1}
\end{equation}
steps.

On WikiHow, MDLM is trained for $\approx 29.8$ epochs at each context length,
with a global batch of 128: $S_{\text{MDLM}} = 28{,}650$ steps at $L = 512$
(123{,}046 training examples, 961.3 steps per epoch) and
$S_{\text{MDLM}} = 41{,}137$ steps at $L = 1024$ (176{,}696 training examples,
1{,}380.4 steps per epoch). Table~\ref{tab:step-matched} gives the resulting
budgets.

\begin{table}[t]
  \centering
  \caption{Step-matched training budgets on WikiHow at context lengths 512 and
  1024. $g_1$ is the fraction of tokens at level~1. Under the summary rule,
  level~1 is the article summary. Under the count rule ($\tau{=}1$), level~1
  is any token occurring at most once in its own sample.}
  \label{tab:step-matched}
  \begin{tabular}{lcccc}
    \toprule
    Model & $g_1$ & $\rho = 1 - g_1/3$ & Steps & Extra steps (epochs) \\
    \midrule
    \multicolumn{5}{l}{\textit{$L = 512$}} \\
    MDLM                 & --      & 1.000 & 28{,}650 & -- \\
    HMDLM (summary)      & 22.34\% & 0.926 & 30{,}955 & $+2{,}305$ ($+2.40$) \\
    HMDLM ($\tau{=}1$)   & 30.82\% & 0.897 & 31{,}930 & $+3{,}280$ ($+3.41$) \\
    \midrule
    \multicolumn{5}{l}{\textit{$L = 1024$}} \\
    MDLM                 & --      & 1.000 & 41{,}137 & -- \\
    HMDLM (summary)      & 16.41\% & 0.945 & 43{,}517 & $+2{,}380$ ($+1.72$) \\
    HMDLM ($\tau{=}1$)   & 27.12\% & 0.910 & 45{,}225 & $+4{,}088$ ($+2.96$) \\
    \bottomrule
  \end{tabular}
\end{table}

\section{Per-Level Schedules and the Induced Reveal Order}
\label{app:ordering}

Levels are indexed so that a higher level is masked more slowly. Formally, for
any two levels $k < k'$ we require
\begin{equation}
  \alpha_{k}(t) \;\le\; \alpha_{k'}(t) \quad \text{for all } t,
  \label{eq:alpha-order}
\end{equation}
with strict inequality at every step at which level $k'$ is active, i.e.
$\alpha_{k'}(t) < 1$. Since $\alpha_k(0) = 1$ and $\alpha_k(T) = 0$ for every
$k$, the inequality is necessarily an equality at the two endpoints. Fewer
level-$k'$ positions are therefore masked at any interior point of the forward
process.

The reverse dynamics are governed by the per-step reveal probability
$\lambda_k(t)$, and \eqref{eq:alpha-order} does \emph{not} by itself imply an
ordering of the $\lambda_k$. We therefore state the property we use directly:
for $k < k'$,
\begin{equation}
  \lambda_{k}(t) \;\le\; \lambda_{k'}(t)
  \qquad \text{whenever } \alpha_{k'}(t) < 1 .
  \label{eq:lambda-order}
\end{equation}
Outside that interval, $\alpha_{k'}(t) = 1$, no level-$k'$ position is masked
and $\lambda_{k'}(t) = 0$; the ordering is vacuous there, since there is
nothing left at level $k'$ to reveal. Both schedules used in our experiments
satisfy \eqref{eq:lambda-order}, as the continuous-time hazards
$\lambda'_k(t) = -\alpha'_k(t)/(1-\alpha_k(t))$ show:
\begin{itemize}[nosep,leftmargin=*]
  \item \textbf{Log-linear with per-level offset} \eqref{eq:alpha_k}, used on
    LM1B: $\lambda'_k(t) = 1/(t - \delta_k)$ for $t > \delta_k$ and $0$
    otherwise. For $\delta_k < \delta_{k'}$ and $t > \delta_{k'}$ we have
    $t - \delta_k > t - \delta_{k'} > 0$, hence
    $\lambda'_k(t) < \lambda'_{k'}(t)$.
  \item \textbf{Power schedule} $\alpha_k(t) = 1 - (1-\varepsilon)\,t^{\omega_k}$,
    used on WikiHow: $\lambda'_k(t) = \omega_k / t$, so the ordering holds
    exactly when $\omega_k < \omega_{k'}$. We use $\omega = (1.0, 2.0)$, and
    $\omega_k = 1$ recovers the log-linear schedule of \mdlm{}.
\end{itemize}
The discrete $\lambda_k(t)$ inherit the ordering on the step grids used in our
experiments ($T = 256$ on both LM1B, and WikiHow): the only
steps at which $\lambda_{0}(t) > \lambda_{1}(t)$ are the $64$ LM1B steps with
$t \le \delta_1$, where level $1$ is inactive and $\lambda_1(t) = 0$.

Consequently, while level $k'$ is still being revealed, a position that Stage~I
predicts to be at level $k'$ receives a larger reveal mass
$\lambda_{k'}(t)\,g^i_{k'}(t)$ in \eqref{eq:reverse_model} and is revealed
earlier, which is the soft generation order \hmdlm{} is designed to induce.

\section{Perplexity: continuous-time ELBO weight}
\label{sec:ppl-weight}

We evaluate test-set perplexity under the continuous-time limit of the
HMDLM ELBO.  For a single hierarchy level $k$, the masking schedule is
linear with offset $\delta_k$:
\begin{equation}
  \alpha_k(t) \;=\;
  \begin{cases}
    1 & t \le \delta_k,\\[2pt]
    1 - (1-\varepsilon)\,\dfrac{t - \delta_k}{1 - \delta_k}
      & \delta_k < t \le 1,
  \end{cases}
  \qquad
 \delta_0=0,  \delta_1 = 0.25,
  \label{eq:alpha-k}
\end{equation}
where $\alpha_k(t)$ is the survival (clean) probability of a level-$k$
token at time $t$.

\paragraph{Per-level continuous-time weight.}
The standard absorbing-state ELBO weight is
\begin{equation}
  \lambda'_k(t)
  \;=\;
  -\,\frac{\alpha_k'(t)}{1-\alpha_k(t)},
  \label{eq:elbo-weight-general}
\end{equation}
For the schedule in~\eqref{eq:alpha-k}, on the active interval $t > \delta_k$,
\begin{align}
  \alpha_k'(t)        \;&=\; -\,\frac{1-\varepsilon}{1-\delta_k},
  &
  1-\alpha_k(t)       \;&=\;(1-\varepsilon)\; \frac{t-\delta_k}{1-\delta_k}.
  \label{eq:alphak-derivs}
\end{align}
Substituting into~\eqref{eq:elbo-weight-general},
\begin{equation}
  w_k(t)
  \;=\;
  \frac{(1-\varepsilon)/(1-\delta_k)}
       {(1-\varepsilon)\,(t-\delta_k)/(1-\delta_k)}
  \;=\;
  \frac{1}{t-\delta_k}.
  \label{eq:weight-cancel}
\end{equation}
For
$t \le \delta_k$ the level is fully unmasked
($1-\alpha_k(t) = 0$) and contributes no loss; we therefore extend $\lambda'_k$
by zero in that regime:
\begin{equation}
  \lambda'_k(t) \;\coloneqq\;
  \begin{cases}
    \dfrac{1}{t-\delta_k} & t > \delta_k,\\[6pt]
    0 & t \le \delta_k.
  \end{cases}
  \label{eq:lambda-k}
\end{equation}
$\delta_0 = 0$ recovers the MDLM weight $1/t$ used
by~\citet{sahoo2024simple}, so~\eqref{eq:lambda-k} is a
strict generalisation.

\paragraph{Posterior-weighted per-token weight.}
Since at evaluation, the gold hierarchy is not available, at position $i$, we replace the gold-hierarchy weight
$\lambda'_{h_0^i}(t)$ with its expectation $r_i(t)$ under the model's posterior over
hierarchy levels at time $t$:
\begin{equation}
  r_i(t)
  \;=\;
  \sum_{k=0}^{K-1}
    [h_\theta^i(x_t, h_t, t)]_h
    \lambda'_k(t),
  \label{eq:pi-i}
\end{equation}
where $h_\theta^i$ is the Pass-1 hierarchy distribution at position $i$.

\paragraph{Token-level perplexity.}
Let $N$ denote the total number of tokens (non-padding) in the test set,
and let $\mathbf{x}_t,\mathbf{h}_t$ be sampled by the joint forward
process at time $t \sim \mathcal{U}(\varepsilon, 1)$, with $\varepsilon = 10^{-3}$.  The per-token
negative log-likelihood is
\begin{equation}
  \mathrm{NLL}_{\mathrm{tok}}
  \;=\;
  \frac{1}{N}\;
  \mathbb{E}_{t,\;\mathbf{x}_t}\!\!\left[
    \sum_{i:\,x_t^i = M}
      r_i(t)\;\bigl(-\log p_\theta(x_0^i \mid \mathbf{x}_t)\bigr)
  \right],
  \label{eq:nll-tok}
\end{equation}
and the reported test-set perplexity is
\begin{equation}
  \mathrm{PPL}
  \;=\;
  \exp\!\bigl(\mathrm{NLL}_{\mathrm{tok}}\bigr).
  \label{eq:ppl}
\end{equation}

\paragraph{Numerical handling of $t \!\to\! \delta_k^{+}$.}
At $t = \delta_k$ is bounded in code by a
denominator floor:
\begin{equation*}
  \lambda_k(t)
  \;\approx\;
  \frac{1}{\max(t - \delta_k,\;\eta)},
  \qquad
  \eta = 10^{-8},
\end{equation*}
which leaves the weight unchanged everywhere except in a vanishingly
small neighbourhood of $\delta_k$ and prevents single-sample gradient
spikes from the Monte-Carlo estimator.

\section{Architecture Details and Qualitative analysis on sudoku}
\label{sec:sudoku_qual}

\subsection{Architecture Details}
For the 6M MGDM and HMDLM models, we use a compact 3-layer GPT-2 Transformer (~5.5M parameters) with hidden size 384, 12 attention heads of dimension 32, a 1536-dimensional GELU feed-forward layer, learned absolute position embeddings, and a 31-token character-level vocabulary whose output head shares weights with the input embedding and bidirectional attention. The hierarchical variant (HMDLM) adds only a level embedding and a level-prediction head, about 5K extra parameters.

For the 25M MDLM and HMDLM models, we use a 12-layer diffusion transformer (DiT, ~25.0M parameters) with hidden size 384, 6 attention heads of dimension 64, a 1536-dimensional GELU feed-forward layer, rotary position embeddings, and bidirectional attention over a fixed 81-token sequence (one token per cell). Both the models are trained from scratch. The hierarchical variant (HMDLM, K=6 levels) adds a 7x384 level embedding (6 levels plus a [MASK] level), which is summed with the token embedding, and a level-prediction head that has the same form as the token output head. Together they add about 104K parameters (25.10M versus 24.99M).

\subsection{Qualitative Analysis}
For the sudoku 25M hmdlm and mdlm models - we identified 287 puzzles where MDLM failed to solve across all three seeds but HMDLM solved correctly across all three. We then performed 8 additional inference runs per model on these puzzles.
 We pick some of these examples and trace their unmasking order to show how HMDLM helped in getting the right answer:

Example 1: 
\texttt{50607080040000001600000000000200000307005040000090002096}\\
\texttt{0000500030002007000100000}

\begin{figure}[H]
    \centering
    \begin{minipage}[c]{0.8\textwidth}
        \includegraphics[width=\linewidth]{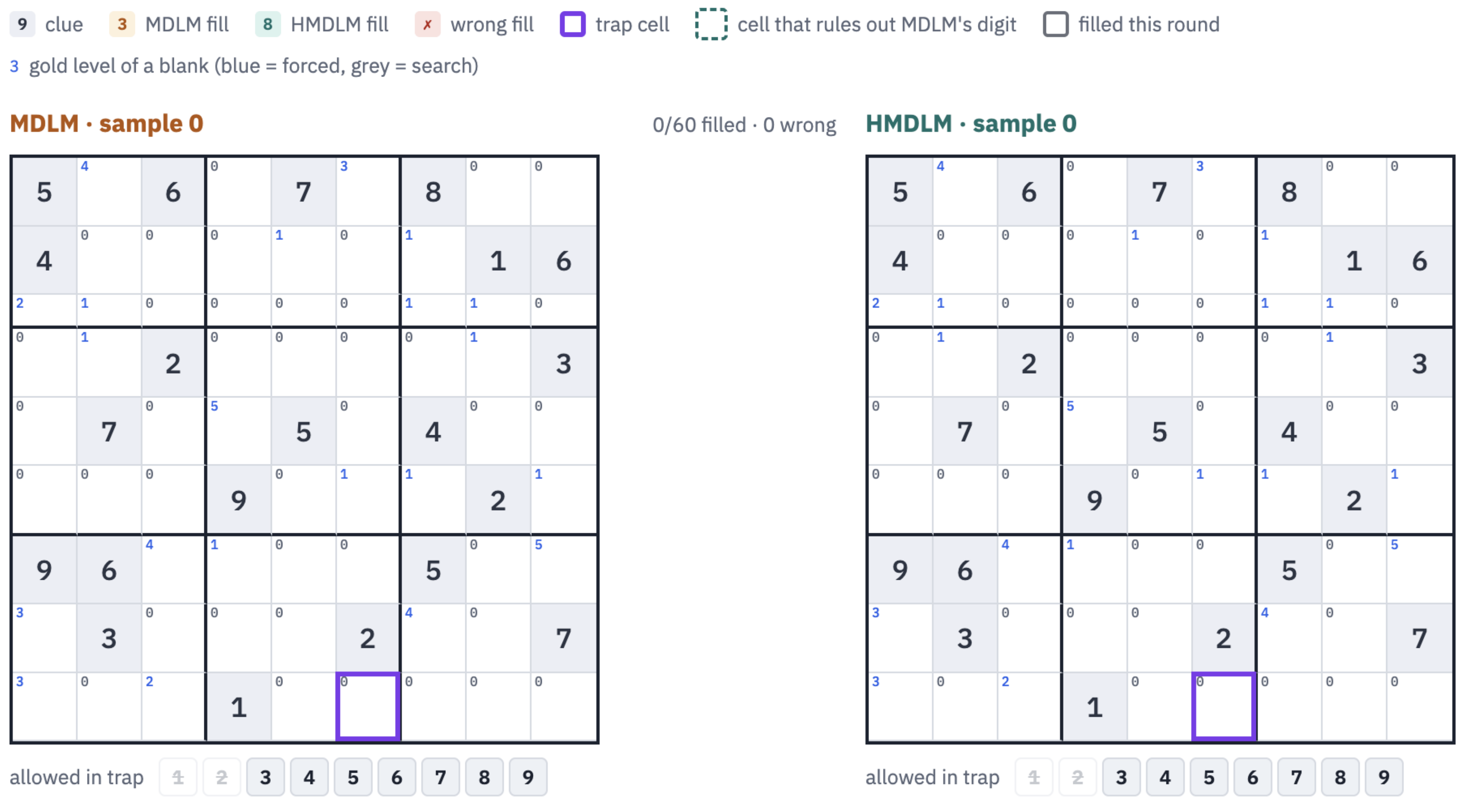}
    \end{minipage}\hfill
    \begin{minipage}[c]{0.20\textwidth}
        \small
        \textbf{Start.} Initial board.
    \end{minipage}
    \caption{Example 1, Start.}
    \label{fig:sudoku_ex1_start}
\end{figure}

\begin{figure}[H]
    \centering
    \begin{minipage}[c]{0.8\textwidth}
        \includegraphics[width=\linewidth]{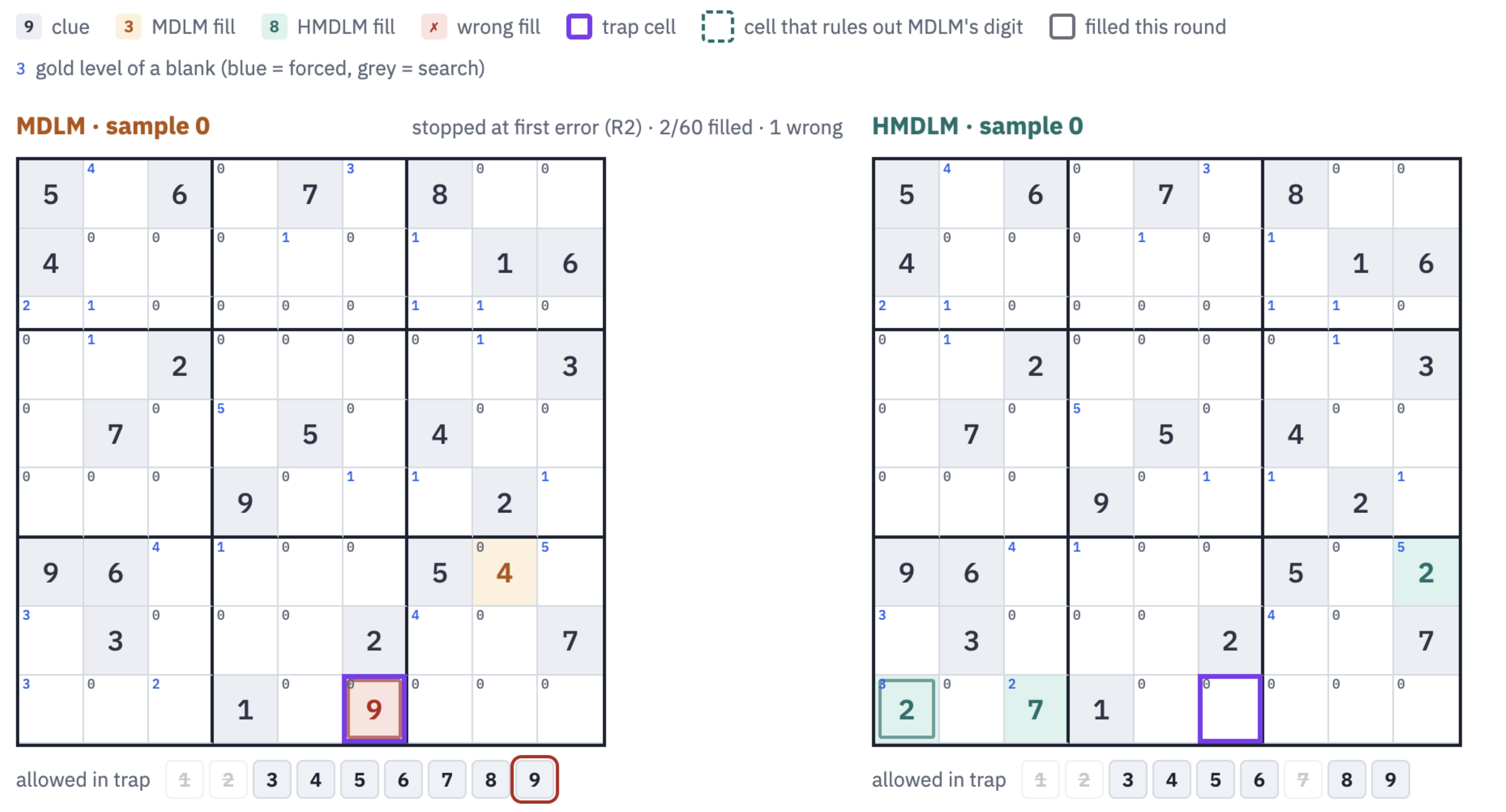}
    \end{minipage}\hfill
    \begin{minipage}[c]{0.20\textwidth}
        \small
        \textbf{Round 2.} MDLM predicts the R9C6 cell and gets it wrong. There are multiple possible values for the cell 3,4,5,6,7,8,9. HMDLM on the other hand, pushes the unmasking of this cell to later rounds.
    \end{minipage}
    \caption{Example 1, round 2.}
    \label{fig:sudoku_ex1_r2}
\end{figure}

\begin{figure}[H]
    \centering
    \begin{minipage}[c]{0.8\textwidth}
        \includegraphics[width=\linewidth]{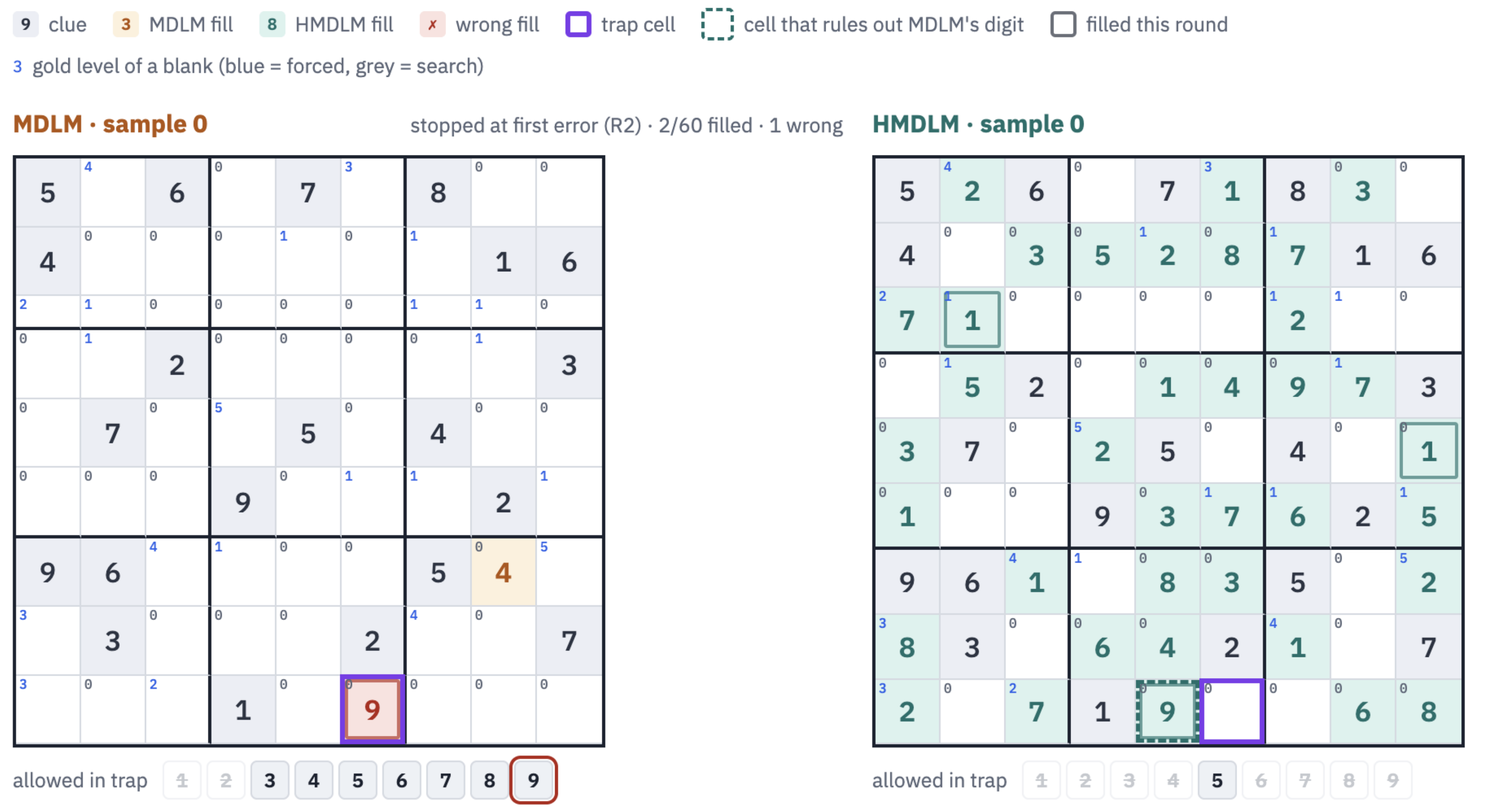}
    \end{minipage}\hfill
    \begin{minipage}[c]{0.20\textwidth}
        \small
        \textbf{Round 25.} HMDLM defers R9C6 and fills other cells first.
    \end{minipage}
    \caption{Example 1, round 25.}
    \label{fig:sudoku_ex1_r25}
\end{figure}

\begin{figure}[H]
    \centering
    \begin{minipage}[c]{0.8\textwidth}
        \includegraphics[width=\linewidth]{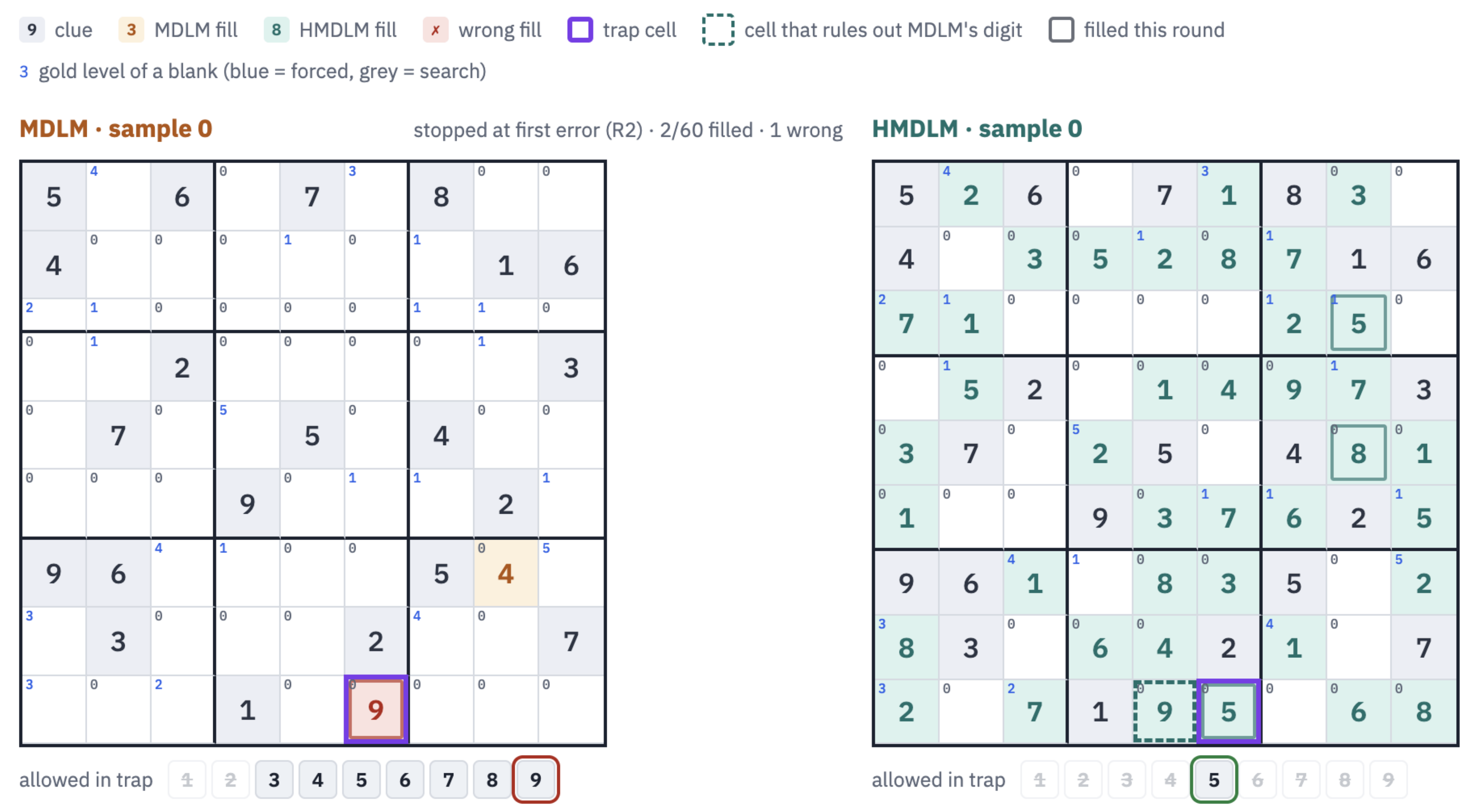}
    \end{minipage}\hfill
    \begin{minipage}[c]{0.20\textwidth}
        \small
        \textbf{Round 26.} Finally, HMDLM gets the R9C6 cell right at round 25, when the only possible value left is 5 and gets it correct.
    \end{minipage}
    \caption{Example 1, round 26.}
    \label{fig:sudoku_ex1_r26}
\end{figure}

Example 2:
\texttt{008100026000807000900000000010008900030000010000450000000309800}\\
\texttt{200000400500020300}

\begin{figure}[H]
    \centering
    \begin{minipage}[c]{0.8\textwidth}
        \includegraphics[width=\linewidth]{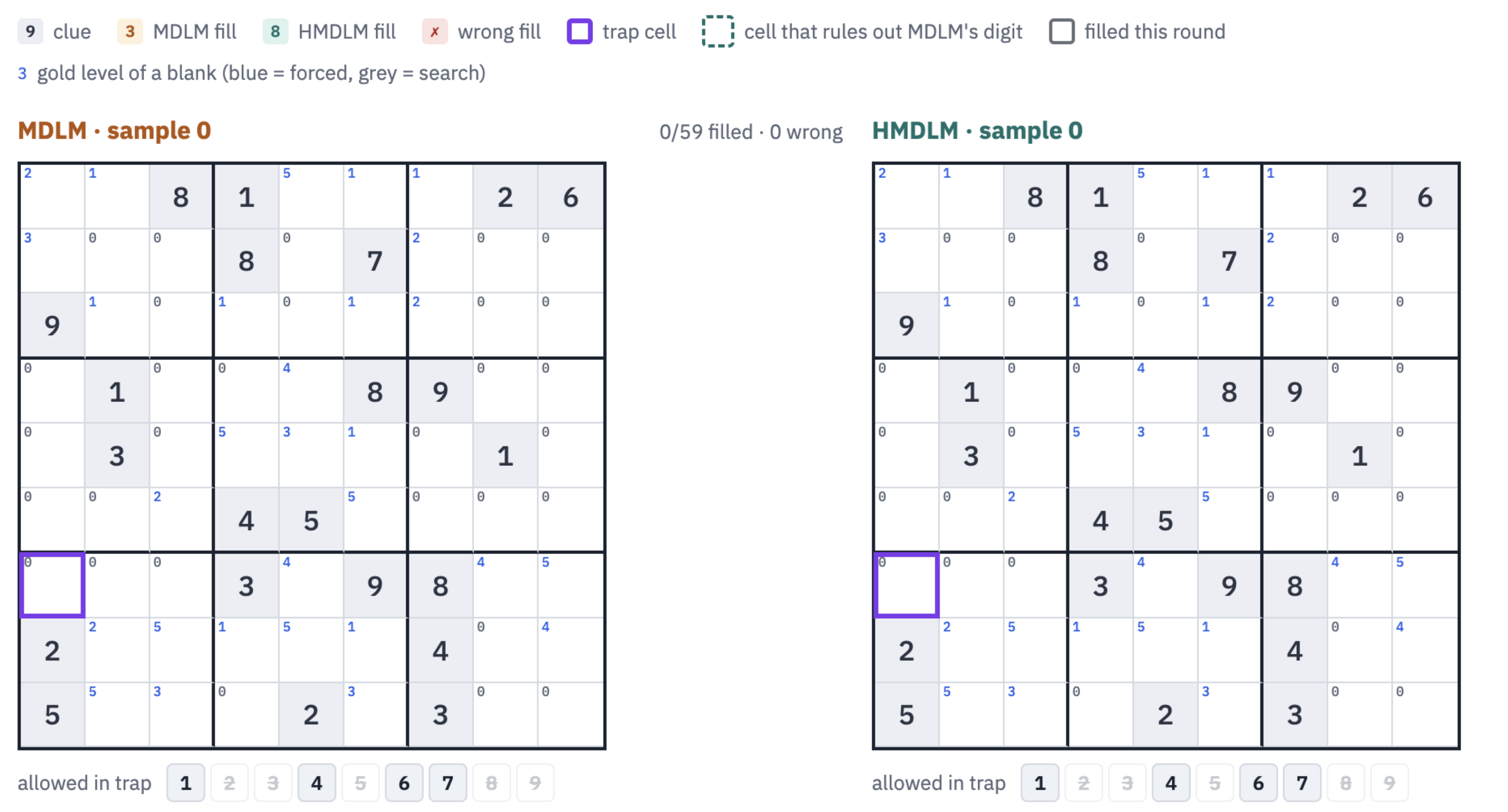}
    \end{minipage}\hfill
    \begin{minipage}[c]{0.20\textwidth}
        \small
        \textbf{Start.} Initial Board.
    \end{minipage}
    \caption{Example 2, Start.}
    \label{fig:sudoku_ex2_start}
\end{figure}

\begin{figure}[H]
    \centering
    \begin{minipage}[c]{0.8\textwidth}
        \includegraphics[width=\linewidth]{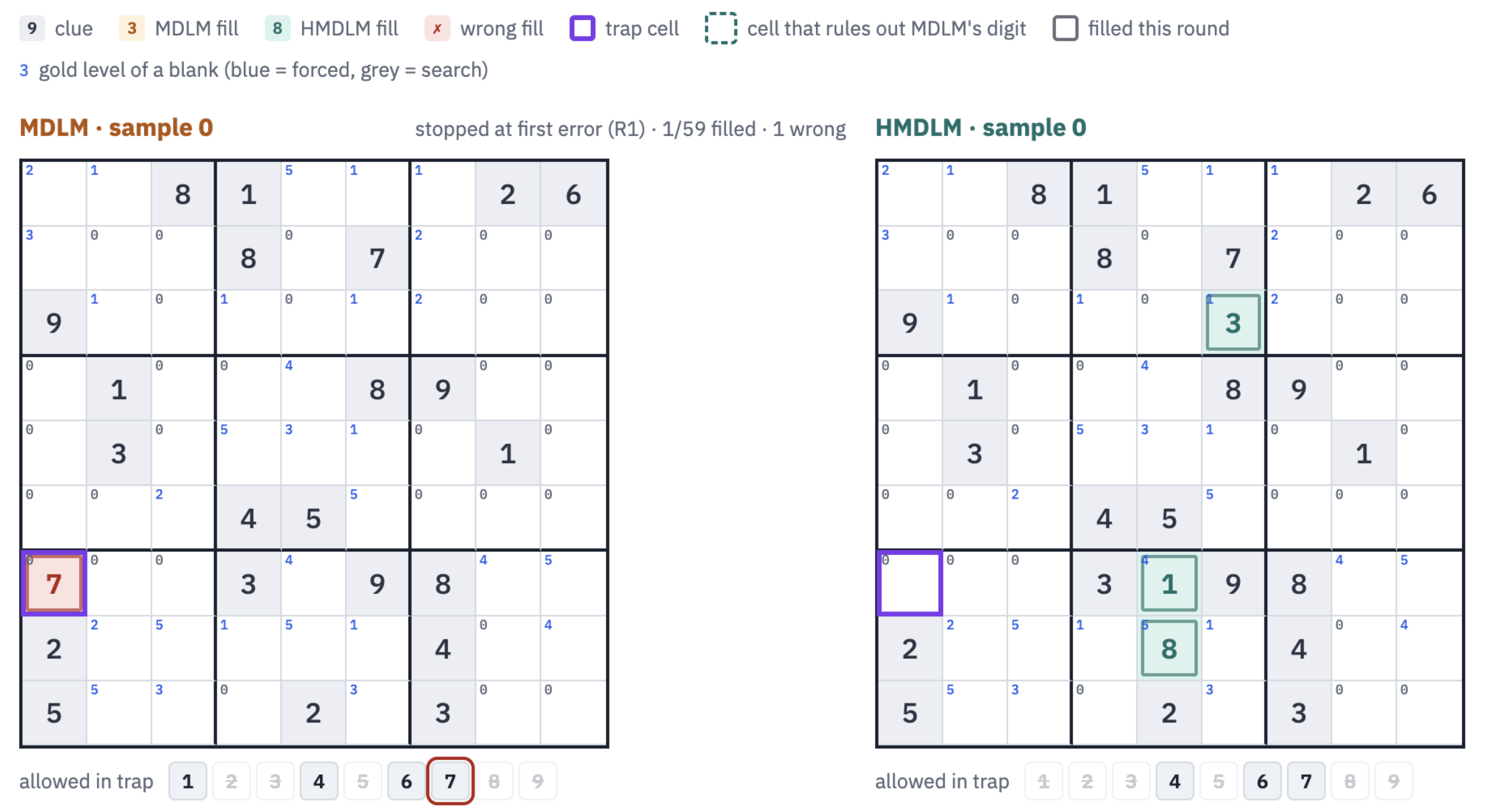}
    \end{minipage}\hfill
    \begin{minipage}[c]{0.20\textwidth}
        \small
        \textbf{Round 1.} MDLM predicts the R7C1 cell and gets it wrong. There are multiple possible values for the cell 1,4,6,7. HMDLM on the other hand, pushes the unmasking of this cell to later rounds.
    \end{minipage}
    \caption{Example 2, round 1.}
    \label{fig:sudoku_ex2_r1}
\end{figure}

\begin{figure}[H]
    \centering
    \begin{minipage}[c]{0.8\textwidth}
        \includegraphics[width=\linewidth]{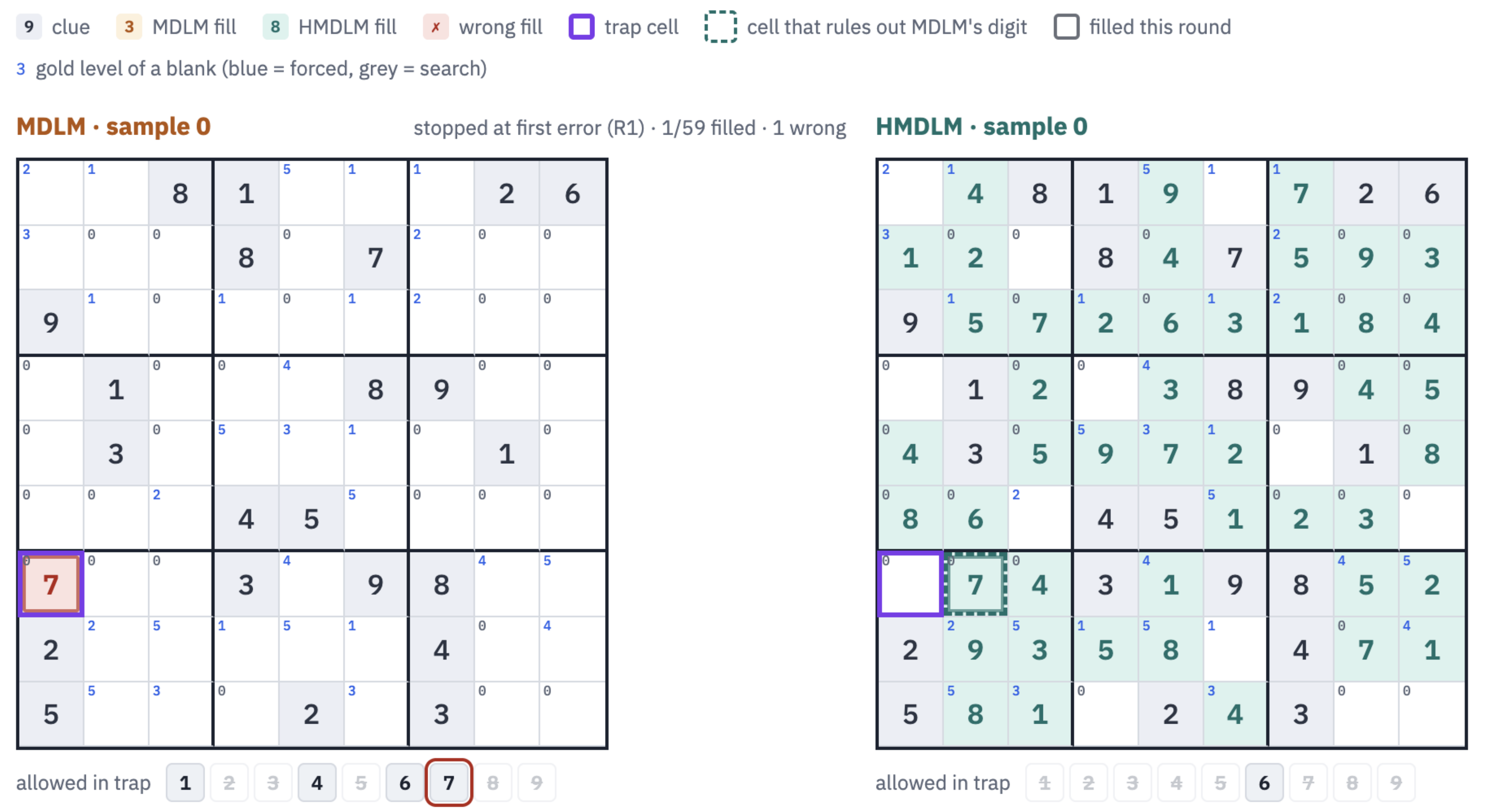}
    \end{minipage}\hfill
    \begin{minipage}[c]{0.20\textwidth}
        \small
        \textbf{Round 29.} HMDLM predicts the neighbouring cells first.
    \end{minipage}
    \caption{Example 2, round 29.}
    \label{fig:sudoku_ex2_r29}
\end{figure}

\begin{figure}[H]
    \centering
    \begin{minipage}[c]{0.8\textwidth}
        \includegraphics[width=\linewidth]{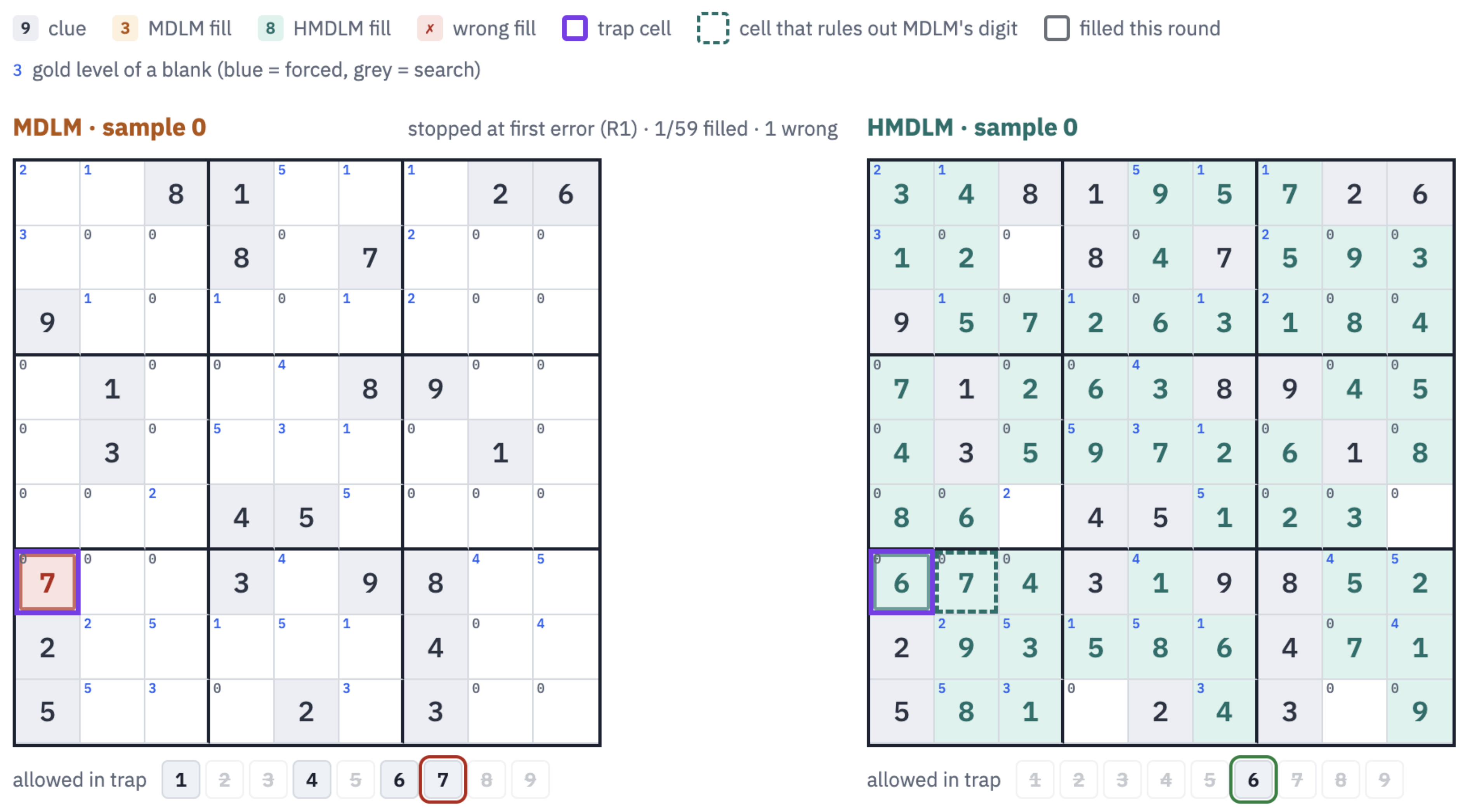}
    \end{minipage}\hfill
    \begin{minipage}[c]{0.20\textwidth}
        \small
        \textbf{Round 36.} HMDLM predicts cell R7C1 correctly.
    \end{minipage}
    \caption{Example 2, round 36.}
    \label{fig:sudoku_ex2_r36}
\end{figure}

Both the samples show the same pattern, MDLM tends to unmask ambiguous cells in the beginning and gets the prediction wrong. On the other hand, HMDLM tends to defer the unmasking of such cells to later rounds until neighbouring cells get filled first. Across 287 puzzles (where HMDLM passed and MDLM failed) with 8 samples per model, we find 131 cases where MDLM's first error is an early commit to a search cell and HMDLM fills that same cell only once a single candidate remains.

\subsection{Cell Accuracy per Hierarchy Level}

\begin{table}[H]
\centering
\small
\setlength{\tabcolsep}{4pt}
\setlength{\abovecaptionskip}{2pt}
\setlength{\belowcaptionskip}{2pt}
\caption{Cell accuracy (\%) by hierarchy level.}
\label{tab:sudoku-cell-levels}
\begin{tabular}{clcc}
\toprule
Level & Cells & MDLM & HMDLM \\
\midrule
5 & givens + round 0        & 99.76 & \textbf{100.00} \\
4 & round 1                 & 99.59 & \textbf{99.99} \\
3 & round 2                 & 99.29 & \textbf{99.98} \\
2 & round 3                 & 98.73 & \textbf{99.92} \\
1 & rounds 4--5             & 96.77 & \textbf{99.45} \\
0 & round 6+ / never forced & 78.78 & \textbf{86.57} \\
\bottomrule
\end{tabular}
\end{table}
Each cell is assigned a level by the constraint-propagation round in which it becomes forced: level 5 cells are givens or forced immediately (round 0), while level 0 cells are forced only after 6+ rounds or never, and so require search. We see that HMDLM improves the accuracy for the most difficult cells (label 0) by up to $8\%$ than MDLM.
\section{Qualitative Analysis on WikiHow}
\label{sec:wikihow_qual}

In this section we look at the generative quality of samples from HMDLM, MDLM, and ADLM models. 

\subsection{Prompt: How to replace the rear shock on an 1150rt}

\paragraph{MDLM}
Find the parts that you need to build in order not to replace the rear shock. diagnose the problem. assemble your latest model. that will take you to points where you want to attach ground cable. shop older models with an antenna kit — the replacement might come with 80mm screw for a zoom lens. when you find your pictures of the 110054ares, youll see an obvious slot on the shelf. disconnect the screws. you may disconnect them above the antenna mast if you connect the kit on the antenna that acts as the zoom lens. tighten up the screws. the wiring actually begins. enjoy wiring it. also make sure that the carbon fiber shaft meets the rest of the parts. that then download the latest version of the wirelesss and chargers. so by now, you will complete your replacement order. youll need to separate the stand like the new shock like your christmas boots from the cables that you use to replace your rear and front shocks. on model ebola suggest that the connection and the handle on disengage. then you should isolate the wires from a stainless steel. set up your stepper mount separately. it identifies the wattage and this one turns on the wire. you will need wiring
\paragraph{ADLM}
Remove the clutch and cut the 17 braided cable from valvages up to about 12th inch into the driver shock. insert the valve valve in a branch of the comparison valve about 1 mm in the bottom lip.  insert in the valve valve with a minimal cussist pressure about 5, 000remp.  push your gently next to a pump station, four thumbs up and the lever about two inches down.  insert one of the bearings into the hard valve, which usually fits into the internal part.  along with pulling on a slow wheel, the vr should be left into the quick valve.  insert a pin, but novel adjustment is an option in the bottom lip.  make sure it stills your heel while pushing the valve, as found in normal mode.  insert in two screws into the vr and tighten the jl band.  begin twisting the traction steel - socket into the rac socket, lid of the driver or shock.  push through the valve flex and, using a cork, or cauterized rubber band, remove the slide from the inside., the traction valve cuts the force of the cussist when the valve is closed.
\paragraph{HMDLM}
Check the components you need to help you replace the rear shock. unscrew the rear bolt to the rear. put the shocks on the ground. remove the shock bolt you have. now you can replace the front shocks. remove the rear bolt with the nut that the rear shock bolt is attached to. with the wheel on the back on the open side, pull. you should pull this with the nut that the front bolt is near it. you should do this in the direction on the rear bolt shown with the bolt. when you do this, you should notice how the pressure is applied. when you do this, you should find that the rear bolt, shock bolt are in place. after this, find that the rear bolt is in place with the front shock bolt. do this a couple of times until you can the find the replacement parts.

\paragraph{Conclusion:}
Among the three samples, HMDLM produces the most faithful response: it stays on the task of replacing a rear shock throughout and refers to relevant components such as shocks, bolts and nuts, although its step order is repetitive and somewhat confused. ADLM maintains a mechanical register but is largely incoherent, with invented terms (e.g., “valvages”, “cussist”) and physically implausible instructions. MDLM drifts off topic almost immediately, moving to antenna kits, zoom lenses and wiring, and never addresses the shock replacement. Overall, HMDLM shows the strongest prompt adherence and topical consistency, ADLM is second, and MDLM is the weakest.

\subsection{Prompt: How to buy a custom fit car cover}

\paragraph{MDLM}
Find a vanity cover if there are none on your list. celebrate who is going to drive. buy the car cover you need. get it framed. label the cover. make sure that it isnt just a side cover but if you still have one, get a frame and create some polo paint. coat your cover with black to make it, drywall post paper or more ply window covers for material covers. get a cheap cover. these will be cheaper than your usual store window cover. since youre planning to sell the car youre going to build you either choose a cover with floor windows. if you do not intend to build it as is, be sure to buy an inexpensive cover. car cover cant go over the years, unless you have a plan to make your best youll always need to make up new windows going forward. tips are useful, so select a cover and not a framing cover as will help sell the idea that it works for that purpose. accessories on the cover cover will free up cover wastage, if you cant find the little scratches in the car you need to wipe them out before youve made your final cover.
\paragraph{ADLM}
, first, make sure the you have the nail links. go down the edge. look at the details on top. put detail here and think some thoughts on if your arms fit correctly.  look toward cushier areas where you would like you get a good fit  like the pieces under the jeep under the convenience button.  frames is fine, but less than works.  where you find where you think you are more fit, good. look over the shoulders. the rest of the seats unless you get at the side of the neck try some chair. find the front cover and try the front wheel. put a few inserts near the peephole. add some glosses or any more products in the top section.  look over the interior of the cover, slip or pop the back.  are you wearing a high top, or just would like to see the final mix out dress options couch pants  did just get too tight for the shoulders  are you fit  do you want the skin soft or tight attention to face, hair, etc. are you sure of movement a little caution.

\paragraph{HMDLM}
Look at the records of most retailers. locate the cover that fits the vehicle youre looking to buy. look for car covers online. locate the stores you can find your custom fit car covers at. look for different types of car covers from depending on what youre buying or looking for specific vehicles. visit a local shop and call and ask them if they will get what you are looking for. pay for or buy custom covers. some stores may keep car bins or keep a box of cars that will fit the type of cover that you can get from the shop. if you want medium - sized vehicles to be as large as you want, you can be able to buy it . however, it is usually not much more than what you want. look for a car type store or a retailer that tailor a car cover to you for a cheaper price.
\paragraph{Conclusion}
HMDLM produces the most relevant response: it stays focused on buying a custom-fit car cover and follows a sensible sequence of steps (checking retailers, searching online, contacting local shops and comparing prices), although a few sentences are garbled (e.g., keep a box of cars). MDLM repeatedly mentions car covers but quickly drifts into unrelated and contradictory instructions such as framing the cover, applying "polo paint" and adding windows. ADLM departs furthest from the prompt, shifting from car covers to garment fitting (shoulders, pants, dress options) with largely incoherent instructions. Overall, HMDLM shows the strongest prompt adherence and coherence, followed by MDLM, with ADLM the weakest.

\section{Prompt for LLM-as-judge}
\label{sec:prompt}

\subsection{LM1B Prompt}
For evaluating the LM1B generated samples under LLM-as-judge, we pass the following prompt:

"""You are an expert evaluator judging two pieces of machine-generated English text. Your task is to determine which text is BETTER in overall quality. You must choose exactly ONE of:

A - Sample A is better

B - Sample B is better

tie - The two samples are approximately equal in quality

There is NO prompt or instruction the text is responding to; these are
standalone generated passages. Evaluate each text holistically, considering:

1. Fluency:
   Is the text grammatical and natural English?

2. Coherence:
   Do the sentences follow each other in a logical, connected way?
   
3. Consistency:
   Does the text stay on a consistent topic without contradictions or
   abrupt, incoherent drift?
   
4. Overall readability:
   Does the passage read like plausible, well-formed human-written text?
   
SCORING RULES:

1. Judge ONLY the two texts.

2. Do NOT assume that either sample is better because it is longer.

3. Do NOT penalize a response merely for being brief or terse.

4. Judge each text independently on its own quality.

5. Do NOT use any information about the source/model that generated the text.

6. The labels A and B are randomly assigned and have no meaning.

7. Choose tie only when the two texts are genuinely comparable in overall quality.

Return ONLY valid JSON:
"winner": "A" | "B" | "tie"
"""

\subsection{WikiHow Prompt}
On similar lines, for running the LLM-as-judge on WikiHow dataset we use the following prompt:

"""You are an expert evaluator judging four responses to the same how-to prompt.

Your task is to determine which ONE response is the BEST for the given prompt.

You must choose exactly ONE of: A, B, C, D.

Evaluate the responses holistically, considering:

1. Adherence:
   Does the response address the specific topic in the prompt, or does it drift?
   
2. Coherence:
   Do the steps follow a logical, connected order?
   
3. Fluency:
   Is the response grammatical and natural English?

4. Overall usefulness:
   Would the response plausibly help someone accomplish what the prompt asks?

SCORING RULES:

1. Judge ONLY the prompt and the four responses.
2. Do NOT assume that a sample is better because it is longer.
3. Do NOT penalize a response merely for being brief or terse if it adequately answers the prompt.
4. Judge each response independently against the prompt.
5. Do NOT compare based on writing style alone; prioritize usefulness and correctness for the prompt.
6. Do NOT use any information about the source/model that generated the responses.
7. The labels A, B, C and D are randomly assigned and have no meaning.
8. If several responses are close, still pick the single best one.

Return ONLY valid JSON:

{
    "winner": "A" | "B" | "C" | "D"
}
"""

\section{Societal Impact of the Model}
\label{sec:societal_impact}
The proposed HMDLM advances masked diffusion language modeling by introducing hierarchical decoding, which improves reasoning and text generation quality. The positive societal impacts include more efficient and accurate language models that can benefit downstream NLP applications such as question answering, text generation, and reasoning tasks. On the negative side, as with all pre-trained language models, HMDLM could potentially be misused to generate misleading or harmful text. 

\section{Computational Resources}
\label{sec:resource}
For language modelling, all experiments were conducted on NVIDIA A100 80GB GPUs. The HMDLM model was trained for approximately four days on 8 GPUs with a batch size of 512, completing 2.5M (~65B tokens) training steps on LM1B dataset.

For WikiHow, All experiments were conducted on NVIDIA A100 80GB GPUs. The HMDLM model was trained for approximately 8 hours on 4 GPUs.

For Sudoku, again we used the same machine and trained on 4GPUs for 4 hours.